\documentclass[a4paper,fleqn]{cas-dc}

\usepackage{capt-of}
\usepackage{booktabs}
\usepackage{multirow}
\usepackage{threeparttable}
\usepackage{caption}

\usepackage[authoryear]{natbib}
\usepackage{float}
\usepackage{CJKutf8}

\usepackage{graphicx}
\usepackage{booktabs, amssymb, adjustbox}
\usepackage{makecell}

\usepackage{pgfplots}

\pgfplotsset{compat=1.18}
\pgfplotsset{colormap={bluered}{rgb(0cm)=(0,0,1); rgb(1cm)=(1,0,0)}}

\usepackage{tikz}
\usepackage[table]{xcolor}

\def\tsc#1{\csdef{#1}{\textsc{\lowercase{#1}}\xspace}}
\tsc{WGM}
\tsc{QE}

\usepackage{amsmath}

\begin{document}
\let\WriteBookmarks\relax
\def\floatpagepagefraction{1}
\def\textpagefraction{.001}
\let\printorcid\relax

\shorttitle{}    

\shortauthors{}  

\title [mode = title]{Tortho-Gaussian++: Suppressing Geometric Drift and Floaters for Robust Gravity-Aligned TDOM Generation in Large-Scale Aerial Scenes}

\author[1]{Xiang Wang}
\credit{Formal analysis}

\author[1]{Xin Wang}
\cormark[1]
\credit{Investigation}

\author[1]{Zongqian Zhan}
\credit{Formal analysis}

\affiliation[1]{organization={School of Geodesy and Geomatics, Wuhan University},
            city={Wuhan},
            postcode={430079}, 
            state={Hubei},
            country={China}}

\cortext[1]{Corresponding author: Xin Wang (xwang@sgg.whu.edu.cn)}


\begin{abstract}
3D Gaussian Splatting (3DGS) has demonstrated remarkable performance in aerial scene reconstruction, motivating its recent extension to true digital orthophoto maps (TDOMs) generation. However, when applied to large-scale aerial scenes, existing 3DGS-based methods still suffer from three critical practical limitations: (1) insufficient overlapping views and poorly conditioned viewing ray geometries, particularly near flight-strip boundaries, leave Gaussians under-constrained along specific directions, resulting in unreliable peripheral reconstructions, geometric drift, and floaters; (2) during densification, the \textit{moment-step misalignment} of the Adam optimizer for newly created Gaussians can amplify each subsequent parameter update, exacerbating the drift and floaters; (3) the orthogonal projection reference for splatting TDOMs from 3DGS is typically from manual rotational correction or camera-distribution priors, which might fail to provide a reliable gravity-aligned reference. To address these issues, we propose Tortho-Gaussian++, a novel 3DGS training framework that suppresses geometric drift and floaters and automatically prunes unreliable peripheral regions, coupled with a robust gravity-aligned TDOM rendering solution. First, we explore geometric support via viewing-ray-based geometric observability analysis, quantitatively characterizing the reliability for each Gaussian and subsequently pruning peripheral regions with particularly weak geometric support. Second, the optimizer state initialization is refined for newly created Gaussians during densification, intrinsically suppressing drift and floaters without external priors. Lastly, a robust gravity direction estimation solution is designed, combining image-derived gravity cues, camera-distribution priors and nadir views, to achieve a gravity-aligned orthogonal projection reference for TDOM rendering. Extensive experiments on multiple large-scale UAV (Unmanned Aerial Vehicle) datasets demonstrate that our Tortho-Gaussian++ outperforms vanilla 3DGS by 0.5--3.5\,dB in average PSNR and yields comparable or superior performance to prior-guided 3DGS methods. Notably, it reduces the training time, VRAM and Gaussian number by 45.5\%, 47.2\%, and 58.6\%, respectively. Qualitative evaluations confirm that our method effectively suppresses drift and floaters, while robust gravity direction estimation reduces tilt-induced facade artifacts in TDOMs. Furthermore, as a prior-free training framework, Tortho-Gaussian++ can be readily integrated into existing 3DGS-based pipelines, greatly enhancing the practicality of 3DGS for TDOM generation. Project Web: \href{https://github.com/IRIS-LAB-whu/Tortho-Gaussian-PlusPlus/}{https://github.com/IRIS-LAB-whu/Tortho-Gaussian-PlusPlus/}
\end{abstract}

\begin{keywords}
3D Gaussian Splatting (3DGS) \sep True Digital Orthophoto Map (TDOM) \sep Geometric Observability Analysis \sep Adam Optimizer \sep Gravity-aligned Rendering \sep
\end{keywords}

\maketitle

\begin{figure*}[!t]
    \centering
    \includegraphics[width=\textwidth]{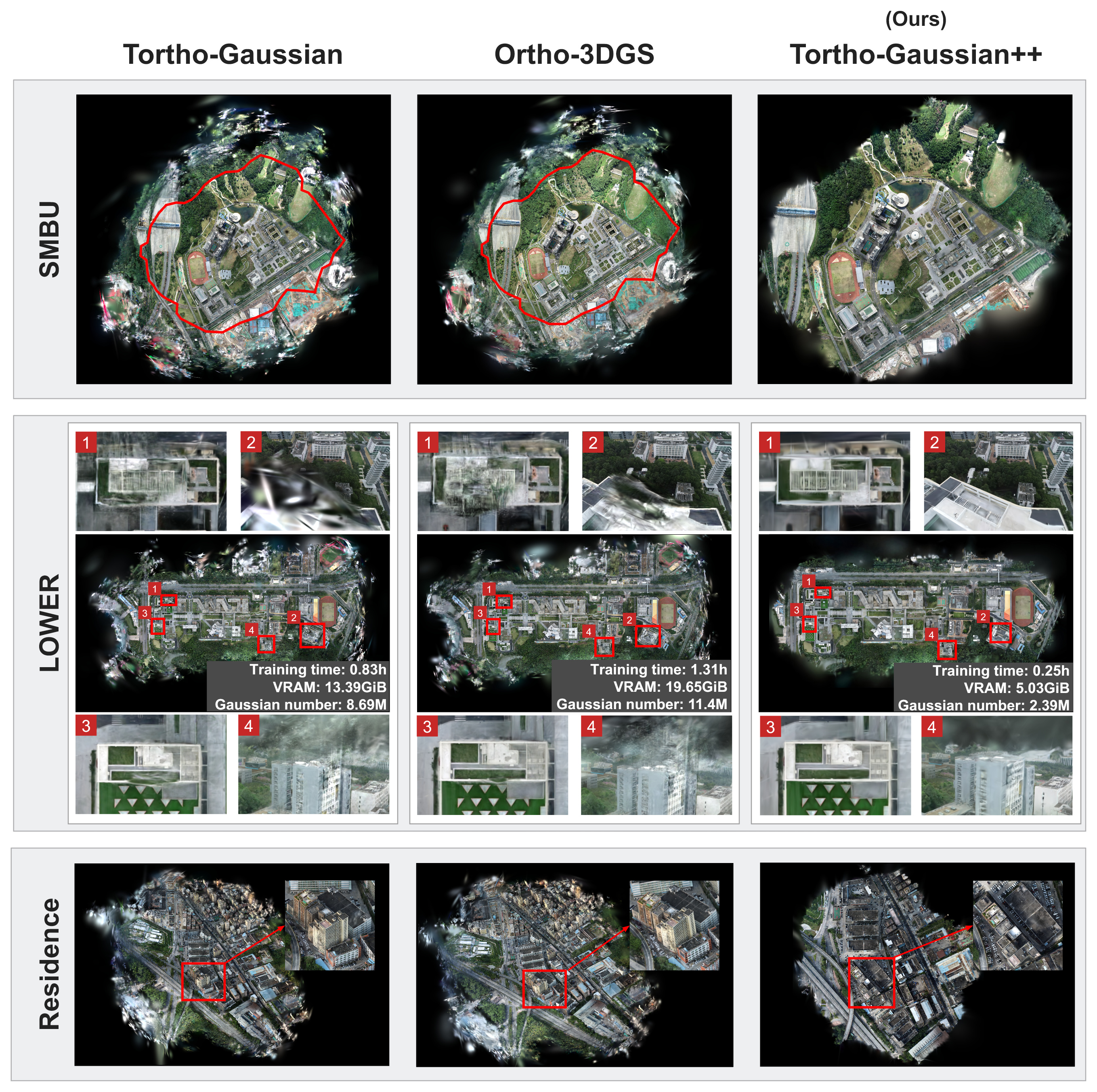}
    \caption{Comparison of Tortho-Gaussian, Ortho-3DGS, and Tortho-Gaussian++ (ours) on three large-scale aerial scenes: \textit{LOWER}, \textit{SMBU} and \textit{Residence}. }
    \label{fig:muscle}
\end{figure*}

\section{Introduction}
\label{sec:introduction}

True Digital Orthophoto Map (TDOM), as a digital foundation, has been widely used in many practical applications, such as digital twins, urban analysis and etc. Traditional photogrammetric pipelines typically employ Structure-from-Motion (SfM) and Multi-View Stereo (MVS) to estimate reliable camera poses and dense point clouds \citep{schonberger2016structure,schonberger2016pixelwise}, followed by digital differential rectification and occlusion detection for TDOM generation \citep{12habib2007new}. Despite achieving high geometric accuracy, these approaches remain computationally demanding for large-scale aerial scenes \citep{li2025ulsr}, and the quality of TDOMs is strictly bounded by the quality of the Digital Surface Model (DSM) \citep{12habib2007new}. Recently, 3D Gaussian Splatting (3DGS) \citep{kerbl20233d} has achieved reconstruction quality comparable to Neural Radiance Field (NeRF) \citep{mildenhall2021nerf}, while offering substantially greater computational efficiency owing to its highly parallelized design and efficient use of GPUs. These capabilities have driven the widespread adoption of 3DGS for TDOM generation via incorporating orthogonal projection \citep{wang2026tortho,yang2025ortho,xu2026tdom,wang2025hygs,wang2026high,yang2026fastpro}, effectively bypassing complex traditional photogrammetric pipelines.

Despite these recent advances, maintaining both high geometric and rendering fidelity remains challenging when applying 3DGS to large-scale aerial scenes \citep{lin2024vastgaussian,liu2024citygaussian,li2025ulsr,zhang2026geogs}. Insufficient overlapping views and poorly conditioned viewing ray geometries can leave the scene geometry insufficiently constrained, leading to severe geometric drift and floaters. To alleviate this problem, existing methods typically incorporate data-driven priors \citep{li2024dngaussian,yang2025ortho,turkulainen2025dn} to guide Gaussian optimization or introduce multi-view geometric consistency constraints \citep{chen2024pgsr,xu2024mvpgs} to improve surface accuracy and suppress reconstruction artifacts. However, the inherent mechanisms underlying geometric drift and floaters remain insufficiently explored. In addition, splatting TDOMs from 3DGS still relies on some manual or heuristic procedures for peripheral cropping and orthogonal directional reference \citep{wang2026tortho,yang2025ortho}. 

Overall, as Fig. \ref{fig:muscle} illustrates, achieving high rendering fidelity and reliable TDOM generation with 3DGS-based methods in large-scale aerial scenes involves the following three practical challenges: 

\textbf{(1) Insufficient overlapping views and poorly conditioned viewing ray geometries.} In regions with limited viewing diversity, the rays intersecting a Gaussian typically span a narrow angular range. For an individual Gaussian, photometric constraints are strong in the plane perpendicular to the viewing ray but much weaker along the ray direction. This directional imbalance creates conditions for Gaussians to move along weakly constrained directions. Gaussians with particularly weak geometric support are often concentrated near flight-strip boundaries, severely degrading overall geometric fidelity. As Fig. \ref{fig:muscle} shows, on the \textit{SMBU} dataset, both the generated TDOM of Tortho-Gaussian \citep{wang2026tortho} and Ortho-3DGS \citep{yang2025ortho} exhibit severely distorted peripheral reconstructions, which are cropped manually. In contrast, Tortho-Gaussian++ automatically prunes unreliable peripheral regions.

\textbf{(2) Optimization-driven geometric drift and floaters.} Although 3DGS adopts the standard Adam optimizer used in deep neural networks (DNNs) \citep{kingma2014adam}, the initialization during densification fails to align the moment history and step count of newly created Gaussians as required by Adam, a phenomenon we term \textit{moment-step misalignment}. This misalignment distorts bias correction and can amplify each subsequent parameter update. Furthermore, under synchronous optimization \citep{ding2026step}, a Gaussian that is invisible in the current viewpoint receives zero gradients but still undergoes parameter updates. Notably, in large-scale UAV scenes, an individual Gaussian is typically invisible in most training views, so such updates occur repeatedly, promoting displacement along weakly constrained directions. While additional geometric constraints can alleviate the resulting errors, \cite{loshchilov2017fixing} point out that the gradient coupling in Adam can destabilize regularization, and these constraints fail to address the underlying \textit{moment-step misalignment}. As shown in Fig. \ref{fig:muscle}, on the \textit{LOWER} dataset, both Tortho-Gaussian and depth-prior-guided Ortho-3DGS yield much more drift and floaters than our Tortho-Gaussian++ in both orthogonal views (1, 3) and perspective views (2, 4).

\textbf{(3) Unreliable orthogonal projection reference.} Existing 3DGS-based TDOM generation methods determine the rendering direction through manual rotational correction \citep{wang2026tortho,xu2026tdom,wang2026high} or camera-distribution priors \citep{yang2025ortho}. Such manual alignment or camera-based heuristics may leave residual errors in the down-vector, resulting in visible building facade artifacts. Their reliability can further deteriorate for datasets containing exclusively oblique views or substantial variations in flight altitude. As Fig. \ref{fig:muscle} illustrates, the \textit{Residence} dataset typifies such challenging scenes, featuring oblique-only views and varying altitudes. Consequently, the rendering directions of Tortho-Gaussian and Ortho-3DGS are severely skewed, whereas Tortho-Gaussian++ preserves a precise orthogonal projection direction.

To address these limitations, we propose Tortho-Gaussian++, a novel 3DGS training framework for high rendering fidelity aerial reconstruction, coupled with a robust gravity-aligned TDOM rendering solution. Our key insight is that poorly conditioned viewing ray geometries create conditions for geometric drift and floaters, while \textit{moment-step misalignment} during densification amplifies the updates that drive them. Moreover, accurate geometry must be paired with a reliable gravity-aligned orthogonal projection reference for TDOM rendering. First, we evaluate geometric support through a viewing-ray-based geometric observability analysis, quantitatively characterizing the reconstruction reliability of each Gaussian and enabling the automatic pruning of peripheral regions with particularly weak geometric support. Second, we refine the optimizer state initialization for newly created Gaussians during densification by transferring accumulated moment estimates from parent Gaussians to their children. This maintains consistency between the moment history and step count, intrinsically suppressing drift and floaters without introducing external priors. Third, we develop a robust gravity direction estimation solution for orthogonal projection reference, which integrates image-derived gravity cues from a pretrained single-image camera calibration network \citep{jin2023perspective}, camera-distribution priors, and nadir views through confidence-based selection and fusion. The resulting down-vector enables gravity-aligned TDOM rendering, significantly reducing tilt-induced facade artifacts in TDOMs.

The main contributions of this work are summarized as follows:

\begin{itemize}

\item We introduce a viewing-ray-based geometric observability analysis that quantitatively characterizes the reliability of each Gaussian in large-scale aerial scenes, enabling automatic pruning of unreliable peripheral regions.

\item We identify moment–step misalignment during densification as a critical source of amplified Gaussian parameter updates, and introduce an optimizer-state inheritance strategy that intrinsically suppresses geometric drift and floaters without relying on external priors.

\item We develop a robust gravity-direction estimation method that combines image-derived gravity cues, camera-distribution priors, and nadir views through confidence-based selection and fusion, enabling gravity-aligned TDOM rendering with reduced tilt-induced facade artifacts.

\end{itemize}

\section{Related Work}
\label{sec:related_work}
This section reviews relevant literature in three key areas: 3D Gaussian Splatting for large-scale scenes, novel view synthesis for TDOM generation, and gravity direction estimation for orthogonal projection.

\subsection{3D Gaussian Splatting for Large-Scale Scene}

3DGS combines explicit scene representations with differentiable rasterization, enabling efficient training and real-time rendering \citep{kerbl20233d}. However, its application to large-scale aerial scenes faces two primary challenges: the high computational and memory costs of massive scenes \citep{lin2024vastgaussian} and limited geometry accuracy under poorly conditioned viewing ray geometries \citep{zhang2026geogs}.

Regarding scalability, global optimization requires maintaining numerous Gaussians and their optimizer states, even though only a fraction contributes to each training view. Processing high-resolution observations further increases rendering costs, motivating scene decomposition and structured training. Spatial partitioning, exemplified by VastGaussian \citep{lin2024vastgaussian}, distributes points and images among cells for parallel optimization. Hierarchical organization in CityGaussian \citep{liu2024citygaussian} combines a coarse global model with local block refinement and level-of-detail representations. Complementing these spatial strategies, Horizon-GS \citep{jiang2025horizon} organizes aerial-to-ground training to accommodate substantial viewpoint differences. While these approaches improve scalability and data utilization, computational and organizational improvements do not directly guarantee reliable local geometry.

Regarding geometric accuracy, photometric consistency does not uniquely constrain surfaces, particularly where limited viewing diversity leaves depth ambiguous. 3D Gaussians can further permit visually similar solutions with different spatial extents, weakening the correspondence between appearance and physical surfaces \citep{huang20242d}. Existing approaches address these issues through four paradigms. \textit{Geometric representation refinement} promotes surface-aligned primitives; for instance, 2DGS \citep{huang20242d} replaces volumetric Gaussians with oriented disks, whereas PGSR \citep{chen2024pgsr} uses flattened primitives and plane-based depth estimation, effectively yielding view-consistent surfaces. \textit{Geometric regularization} imposes consistency on reconstructed geometry; PGSR combines multi-view photometric and geometric constraints, ULSR-GS \citep{li2025ulsr} enforces depth and normal consistency, and GaussianCraft \citep{xiang2026gaussiancraft} introduces gradient-norm regularization to suppress perturbations while preserving fine structures. \textit{Prior guidance} supplies additional geometric information; FSGS \citep{zhu2024fsgs} and DNGaussian \citep{li2024dngaussian} leverage monocular depth guidance, while GeoGS \citep{zhang2026geogs} and Ortho-3DGS \citep{yang2025ortho} incorporate urban model priors or depth guidance to constrain aerial scene geometry. \textit{Densification refinement} adjusts Gaussian generation. AbsGS \citep{ye2024absgs} mitigates gradient cancellation to recover fine details, while Pixel-GS \citep{zhang2024pixel} accounts for pixel coverage in gradient aggregation and camera distance in gradient scaling, thereby improving detail while reducing nearby floaters. For UAV orthophotos, DS-ADC \citep{wang2026high} further introduces fixed multi-view neighborhoods and edge-aware consistency to direct densification toward reliable structures.

Despite these extensive modifications to scene representations or extra imposed supervision, existing paradigms do not explicitly characterize how viewing geometry constrains individual Gaussians, nor do they explain how newly created Gaussians are driven into drift and floaters. Optimization-focused studies provide a complementary perspective. 3DGS-MCMC \citep{kheradmand20243d} formulates Gaussian optimization via stochastic sampling and state transitions, while Sparse Adam in Taming 3DGS \citep{mallick2024taming} restricts updates to active Gaussians, thereby enabling asynchronous optimization. Recent studies on decoupled optimization in 3DGS \citep{kingma2014adam} further show that update-step coupling and moment evolution can cause invisible Gaussians to receive updates and undergo moment rescaling under standard Adam. Together, these findings suggest that optimizer-state dynamics may contribute to unstable Gaussian parameter updates. However, the \textit{moment-step misalignment} during densification—specifically, the misalignment between moment initialization and inherited optimization steps for newly created Gaussians—and how it intrinsically amplifies parameter updates, remain unexplored.

\subsection{Novel View Synthesis for TDOM Generation}

Traditional TDOM generation uses Digital Surface Models (DSMs) for relief correction and occlusion handling \citep{amhar1998generation}. Visibility analysis identifies valid observations and allows occluded areas to be resolved using redundant observations from overlapping imagery \citep{12habib2007new}. Consequently, the quality of the resulting orthophotos is strictly bounded by the accuracy of the underlying surface geometry and visibility estimation.

Novel view synthesis approaches instead generate TDOMs directly from learned scene representations, bypassing complex photogrammetric pipelines. NeRF-based methods integrate density and radiance along parallel rays via orthogonal volume rendering \citep{yue2025nerfortho,chen2025ortho,mildenhall2021nerf,qu2024satellite}. Explicit 3D Gaussian methods generate TDOMs by modifying rasterization and replacing the original perspective projection matrix with an orthogonal matrix. In current NVS-based TDOM generation, the \textit{orthogonal projection direction} is typically determined via two strategies: manual rotational correction \citep{wang2026tortho,xu2026tdom,wang2026high,yue2025nerfortho} or camera-distribution priors \citep{yang2025ortho}. The former relies on manually adjusting a selected nadir camera or applying Manhattan alignment, while the latter exploits the prior of constant UAV flight altitude to establish a plane approximately perpendicular to gravity. Manual alignment depends on subjective interactive judgment, whereas camera-plane priors assume that the acquisition trajectory provides a reliable vertical reference. When the dataset lacks nadir cameras or exhibits substantial flight-altitude variations, the agreement between the selected projection direction and the true gravity direction requires further verification.

In addition to the projection direction, NVS-based TDOM generation typically requires \textit{manual cropping of peripheral regions} after orthogonal rendering to remove distorted areas. This introduces additional manual intervention and heuristic thresholds, which may inadvertently remove peripheral regions that could otherwise be reconstructed with acceptable geometric support, or retain peripheral regions that suffer from poor reconstruction quality. This limitation highlights the necessity for an automatic, geometry-aware mechanism to quantitatively characterize the reliability for each Gaussian and prune unreliable peripheral regions.

\subsection{Gravity Direction Estimation for Orthogonal Projection}

For NVS-based TDOM generation, ensuring consistency between the rendering orientation and the true gravity direction is critical, yet it remains insufficiently discussed in the context of large-scale aerial scenes. Determining this orientation can draw on structural geometry in images or learned single-image orientation cues.

Vanishing-point-based methods infer orientation from projections of parallel scene lines. \cite{tardif2009non} estimate vanishing points through robust line grouping, while CTRL-C \citep{lee2021ctrl} combines image and line features in a transformer for camera calibration. However, their reliability heavily depends on informative line observations and suitable scene structure (e.g., the Manhattan world assumption), which are often limited or ambiguous in weakly structured natural environments or complex urban scenes.

Learned single-image methods exploit broader visual evidence to estimate orientation. \cite{hold2018perceptual} directly estimate camera orientation and field of view. GeoCalib \citep{veicht2024geocalib} combines learned fields with geometric optimization, supporting uncertainty estimation and joint calibration across images. Perspective Fields \citep{jin2023perspective} predicts dense up-vector and latitude fields, describing projected vertical directions and viewing-ray elevations relative to the horizontal plane. For aerial imagery, AGE \citep{zhu2026aerial} fuses DINOv2 features with ray fields derived from known camera intrinsics to improve gravity direction estimation across diverse viewing configurations.

\section{Preliminaries}
\label{sec:preliminaries}

\subsection{3D Gaussian Splatting}

3DGS \citep{kerbl20233d} explicitly represents a 3D scene as a set of 3D ellipsoidal Gaussians initialized from sparse SfM points. Each Gaussian is parameterized by its position $\boldsymbol{\mu}_n$, anisotropic covariance $\boldsymbol{\Sigma}_n$, opacity ${\alpha}_n$, and spherical harmonic (SH) coefficients encoding view-dependent color. The 3D Gaussian is defined as:
\begin{equation}
G_n(\mathbf{x}) =
\exp\left[
-\frac{1}{2}
(\mathbf{x}-\boldsymbol{\mu}_n)^{\top}
\boldsymbol{\Sigma}_n^{-1}
(\mathbf{x}-\boldsymbol{\mu}_n)
\right]
\label{eq:gaussian}
\end{equation}
This full 3D covariance matrix $\boldsymbol{\Sigma}_n$ is parameterized by a rotation $\mathbf{R}_n$ and a scaling $\mathbf{S}_n$:
\begin{equation}
\boldsymbol{\Sigma}_n =
\mathbf{R}_n
\mathbf{S}_n
\mathbf{S}_n^{\top}
\mathbf{R}_n^{\top}
\label{eq:covariance}
\end{equation}

For novel view rendering, each 3D Gaussian is projected onto the 2D image plane (i.e., splatting). Based on the depth sorting, the projected 2D Gaussians are accumulated via front-to-back ${\alpha}$-blending to estimate the pixel color:
\begin{equation}
C =
\sum_{n=1}^{N}
\mathbf{c}_n \alpha'_n
\prod_{j=1}^{n-1}
(1-\alpha'_j)
\label{eq:color_compositing}
\end{equation}
\begin{equation}
\alpha'_n =
{\alpha}_n \exp\left[
-\frac{1}{2}
(\mathbf{x}-\boldsymbol{\mu}'_n)^{\top}
(\boldsymbol{\Sigma}'_n)^{-1}
(\mathbf{x}-\boldsymbol{\mu}'_n)
\right]
\label{eq:alpha}
\end{equation}
where $\mathbf{c}_n$ is the SH-evaluated color and $\mathbf{x}$ denotes the pixel location in normalized image coordinates. Then the Gaussian parameters are optimized by minimizing the photometric loss:
\begin{equation}
\mathcal{L} =
(1-\lambda)\mathcal{L}_1
+\lambda\mathcal{L}_{\mathrm{D\text{-}SSIM}}
\label{eq:photometric_loss}
\end{equation}
During training, the Adam optimizer \citep{kingma2014adam} is employed to update the Gaussian parameters. However, in vanilla 3DGS with synchronous optimization \citep{ding2026step}, 3D Gaussians that receive zero gradients (e.g., those unobserved by the current training view at a specific iteration) may still undergo parameter updates driven by historical moments, a phenomenon that can exacerbate geometric drift.

\subsection{TDOM Rendering with Orthogonal Projection Reference}
In traditional photogrammetric pipeline, TDOMs are generated by interpolating from a DSM, which is inherently parallel to the local geoid. In contrast, rendering TDOMs from a pre-trained 3DGS field without any geotag prior requires defining an \textit{orthogonal projection direction} that approximates the gravity vector in the real 3D scene. Existing 3DGS-based methods establish this reference using camera information; for example, Tortho-Gaussian \citep{wang2026tortho} selects a camera near the centroid of the camera positions and supports manual rotational correction of its orientation, whereas Ortho-3DGS \citep{yang2025ortho} estimates the reference direction from the normal of a plane fitted to the camera centers.

Given the reference down-vector \(\mathbf d\), a virtual camera is oriented along \(\mathbf d\). Its world-to-camera rotation is constructed as:
\begin{equation}
\mathbf{W}(\mathbf{d}) =
\begin{bmatrix}
\mathbf{e}_1^{\top} \\[4pt]
\mathbf{e}_2^{\top} \\[4pt]
-\mathbf{d}^{\top}
\end{bmatrix}
\end{equation}
where \(\mathbf e_1\) and \(\mathbf e_2\) are orthonormal image-plane axes. Given this projection reference, the Gaussian centers and covariances are transformed onto the normalized image plane as:
\begin{equation}
\boldsymbol{\mu}'_n=
\pi\left(
\mathbf{P}
\begin{bmatrix}
\mathbf{W}(\mathbf{d})\boldsymbol{\mu}_n+\mathbf{t}_c\\
1
\end{bmatrix}
\right)
\end{equation}
\begin{equation}
\boldsymbol{\Sigma}'_n=
\mathbf{J}_n
\mathbf{W}(\mathbf{d})
\boldsymbol{\Sigma}_n
\mathbf{W}(\mathbf{d})^{\top}
\mathbf{J}_n^{\top}
\end{equation}

where \(\mathbf{t}_c=-\mathbf{W}(\mathbf{d})\mathbf{c}_o\) is the translation associated with the virtual camera center \(\mathbf{c}_o\), and \(\mathbf{P}\in\mathbb{R}^{4\times4}\) is the camera-to-clip projection matrix. The operator \(\pi\) performs homogeneous division. The matrix \(\mathbf{J}_n\) is the image-plane projection Jacobian with a zero-padded third row.

To generate TDOMs, the perspective projection used during optimization must be replaced by orthogonal projection. The corresponding projection matrix and Jacobian are given by \citep{zwicker2001ewa}:


\begin{equation}
    \mathbf{P}_{\mathrm{ortho}}=
    \begin{bmatrix}
        \dfrac{2}{r-l} & 0 & 0 & -\dfrac{r+l}{r-l}\\
        0 & \dfrac{2}{t-b} & 0 & -\dfrac{t+b}{t-b}\\
        0 & 0 &
-\dfrac{2}{z_{\mathrm{far}}-z_{\mathrm{near}}} &
-\dfrac{z_{\mathrm{far}}+z_{\mathrm{near}}}
{z_{\mathrm{far}}-z_{\mathrm{near}}}\\
       0 & 0 & 0 & 1
    \end{bmatrix}
\end{equation}

\begin{equation}
\mathbf{J}_{\mathrm{ortho}} =
\begin{bmatrix}
\frac{2}{r-l} & 0 & 0 \\
0 & \frac{2}{t-b} & 0 \\
0 & 0 & 0
\end{bmatrix}
\label{eq:orthographic_jacobian}
\end{equation}

where \(l,r,b,t\) specify the left, right, bottom, and top bounds of the orthogonal view volume, respectively, while \(z_{\mathrm{near}}\) and \(z_{\mathrm{far}}\) denote the near and far clipping distances. Substituting \(\mathbf{P}_{\mathrm{ortho}}\) and \(\mathbf{J}_{\mathrm{ortho}}\) into the above transformations projects the Gaussians along parallel rays aligned with \(\mathbf{d}\) onto an image plane.

\section{Methods}
\label{sec:methods}
\begin{figure*}[!t]
    \centering
    \includegraphics[width=\textwidth]{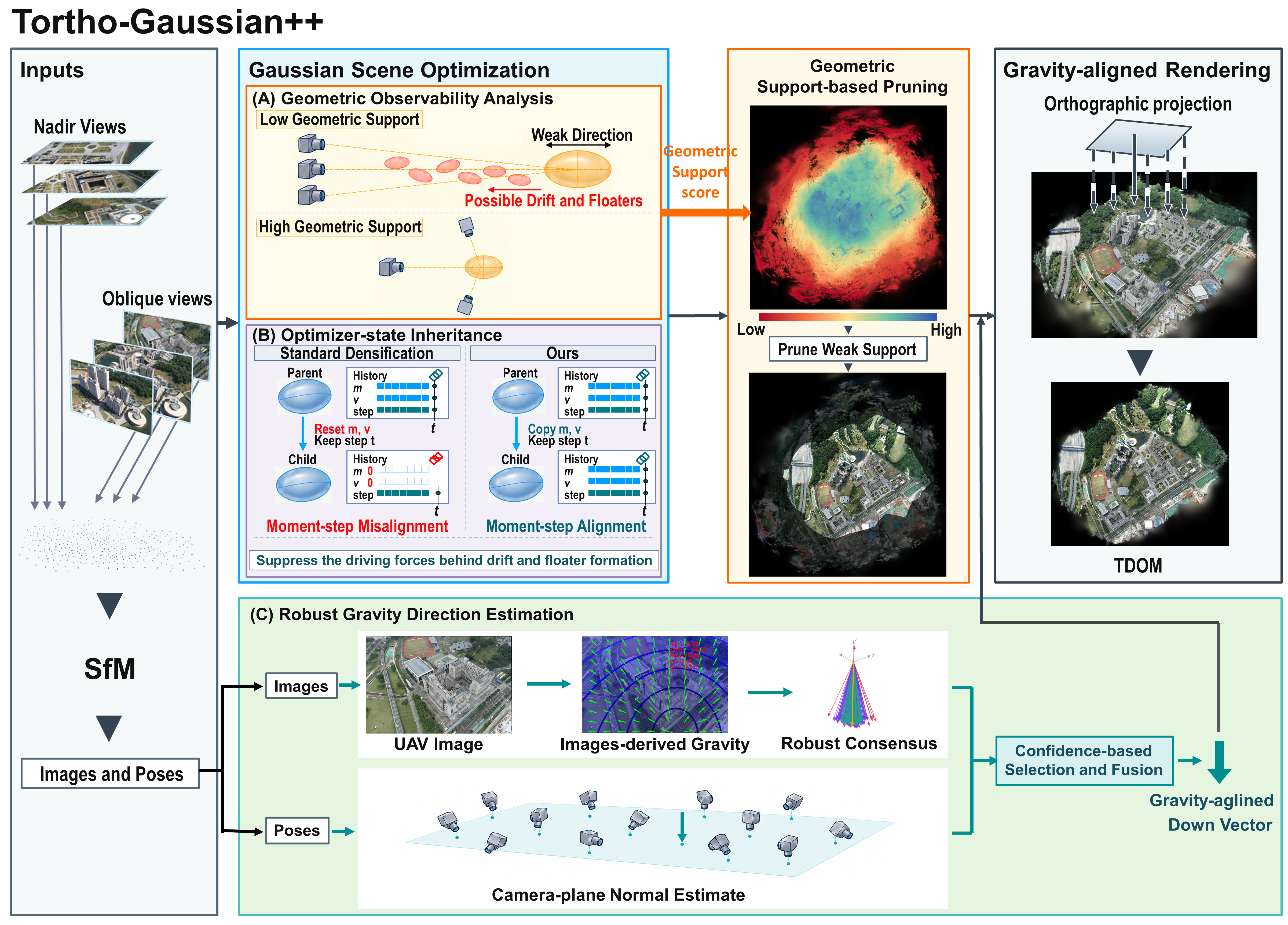}
    \caption{Overview of the proposed Tortho-Gaussian++. (A) Viewing-ray-based geometric observability analysis quantifies the geometric support of each Gaussian during training, enabling automatic pruning of weakly supported Gaussians after training. (B) Optimizer-state inheritance transfers the parent's accumulated moment estimates to newly created Gaussians while retaining the shared optimizer step count, suppressing geometric drift and floaters caused by amplified parameter updates. (C) Robust gravity direction estimation combines image-derived gravity cues, camera-distribution priors, and nadir views through confidence-based selection and fusion. The resulting down-vector establishes a gravity-aligned orthogonal projection reference for rendering the pruned Gaussian scene into a TDOM.}
    \label{fig:framework}
\end{figure*}

As illustrated in Fig. \ref{fig:framework}, we propose Tortho-Gaussian++, a robust 3DGS-based solution for generating TDOMs through automatic pruning of unreliable peripheral regions, suppression of geometric drift and floaters, and gravity-aligned orthogonal rendering. Given 3D sparse points and camera poses from SfM, our pipeline mainly comprises 3DGS training and TDOM rendering. During training, viewing-ray-based geometric observability analysis quantifies the geometric support of each Gaussian (Sec. \ref{sec:3.2}), while optimizer-state inheritance during densification suppresses geometric drift and floaters (Sec. \ref{sec:3.3}). After training, the accumulated support scores guide the automatic pruning of unreliable peripheral 3DGS regions (Sec. \ref{sec:3.2}). For TDOM rendering, a robust gravity direction estimation is employed by combining image-derived gravity cues, camera-distribution priors, and nadir views through confidence-based selection and fusion (Sec. \ref{sec:3.4}). The estimated gravity-aligned down-vector then establishes the orthogonal projection reference for rendering the pruned 3D Gaussian scene into a gravity-aligned TDOM.

\subsection{Viewing-Ray-Based Geometric Observability Analysis}
\label{sec:3.2}
\subsubsection{Directional Constraints from Viewing Rays}
\label{sec:3.2.1}
\begin{figure*}[!t]
    \centering
    \includegraphics[width=\textwidth]{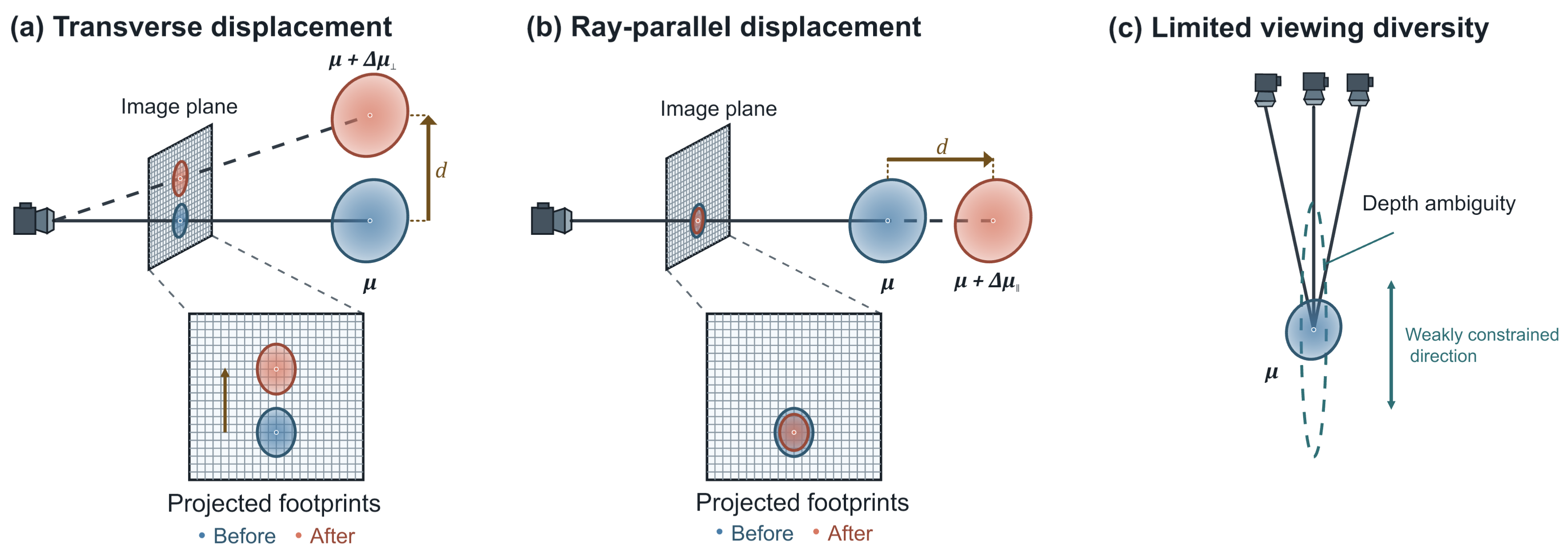}
    \caption{Directional dependence of photometric constraints on Gaussian positions under perspective projection.}
    \label{fig:depth}
\end{figure*}

In standard 3DGS with perspective projection, the photometric constraints on Gaussian positions are highly direction-dependent. As depicted in Fig. \ref{fig:depth} (a), displacing a Gaussian perpendicular to its viewing ray shifts its projected center and redistributes its contributions across neighboring pixels. In regions with informative image content, this displacement directly affects photometric loss, resulting in significant positional gradients. Conversely, as illustrated in Fig. \ref{fig:depth} (b), displacing a Gaussian along the viewing ray preserves the projected center while merely changing the projected footprint scale. This footprint variation can also produce a small but nonzero photometric gradient along the ray. Consequently, the position along the viewing ray is generally much more weakly constrained than the transverse position.

Multiple views can mitigate this ambiguity by providing complementary viewing directions. However, as illustrated in Fig. \ref{fig:depth} (c), a narrow angular range of viewing rays can also leave a common direction weakly constrained despite a large number of observations, which is very common for UAV datasets with high flight altitude. In such poorly conditioned geometries, geometric drift along this direction and the associated formation of floaters cannot be effectively suppressed. This motivates the explicit evaluation of the directional geometric support for each Gaussian from its viewing rays, as developed in Sec. \ref{sec:3.2.2}.

\subsubsection{Multi-view Geometric Support Estimation}
\label{sec:3.2.2}
To yield 3D Gaussians of precise position, existing methods typically alleviate depth ambiguity through depth priors \citep{li2024dngaussian,yang2025ortho} or multi-view geometric consistency constraints \citep{xu2024mvpgs,chen2024pgsr}. However, the directional constraints supplied by viewing rays to individual Gaussians are rarely explicitly quantified. To explore these constraints, we construct a geometric support tensor for each 3D Gaussian, drawing inspiration from the Fisher Information Matrix (FIM) widely used in bundle adjustment \citep{triggs1999bundle} to evaluate parameter observability.

For an individual Gaussian centered at \(\boldsymbol{\mu}_n\) observed by a camera centered at \(\mathbf{c}_j\), the viewing ray from the camera to the Gaussian can be represented by the unit vector:
\begin{equation}
    \mathbf{v}_{nj}
=
\frac{\boldsymbol{\mu}_n-\mathbf{c}_j}
{\rho_{nj}}
\end{equation}
where \(\rho_{nj}=\|\boldsymbol{\mu}_n-\mathbf{c}_j\|\) denotes their L2 distance. Assuming that the camera pose is fixed, differentiating this unit viewing-ray direction with respect to the Gaussian position yields the Jacobian:
\begin{equation}
    \mathbf{B}_{nj}
=
\frac{\partial\mathbf{v}_{nj}}
{\partial\boldsymbol{\mu}_n}
=
\frac{1}{\rho_{nj}}
\left(
\mathbf{I}-\mathbf{v}_{nj}\mathbf{v}_{nj}^{\top}
\right)
\end{equation}

This Jacobian maps a small displacement of the Gaussian center to the corresponding first-order update in viewing direction. Assuming locally isotropic Gaussian noise with variance \(\sigma_{\theta,j}^{2}\) in the tangent plane of the unit sphere, the positional FIM of this linearized bearing-only observation model is given by:

\begin{equation}
    \boldsymbol{\mathcal{I}}_{nj}
=
\frac{1}{\sigma_{\theta,j}^{2}}
\mathbf{B}_{nj}^{\top}\mathbf{B}_{nj}
=
\frac{1}{\sigma_{\theta,j}^{2}\rho_{nj}^{2}}
\left(
\mathbf{I}-\mathbf{v}_{nj}\mathbf{v}_{nj}^{\top}
\right)
\end{equation}

To isolate the angular contribution of the viewing ray, independent of distance and noise scale, we normalize the FIM as:
\[
\mathbf{A}_{nj}
=
\sigma_{\theta,j}^{2}\rho_{nj}^{2}
\boldsymbol{\mathcal{I}}_{nj}
=
\mathbf{I}-\mathbf{v}_{nj}\mathbf{v}_{nj}^{\top}.
\]
The resulting matrix $\mathbf{A}_{nj}$ is an orthogonal projector onto the plane perpendicular to the viewing ray, which corresponds to the 2D subspace effectively constrained by the current input image observation. Its eigenvalues are \(1,1,0\), representing two constrained transverse directions and one unobservable direction in the single-view bearing-only observation model.

For an individual 3D Gaussian, the contributions of different viewing rays are aggregated into a geometric support tensor:
\begin{equation}
    \mathbf{F}_n
=
\sum_{j\in\mathcal{V}_n}
w_{nj}\mathbf{A}_{nj}
\label{eq:w}
\end{equation}
where \(\mathcal{V}_n\) denotes the set of viewing rays and \(w_{nj}\geq 0\) represents their aggregation weights. For any unit vector \(\mathbf d\) representing a possible displacement direction of the Gaussian, the tensor satisfies:
\begin{equation}
\label{dfd}
    \mathbf{d}^{\top}\mathbf{F}_n\mathbf{d}
=
\sum_{j\in\mathcal{V}_n}
w_{nj}
\left[
1-
\left(\mathbf{d}^{\top}\mathbf{v}_{nj}\right)^2
\right]
=
\sum_{j\in\mathcal{V}_n}
w_{nj}\sin^2\theta_{nj}
\end{equation}
where \(\theta_{nj}\) is the angle between \(\mathbf d\) and \(\mathbf v_{nj}\). A displacement direction \(\mathbf d\) nearly parallel to a viewing ray receives minimal geometric support from that view, whereas a direction nearly perpendicular to the ray receives stronger support.

\begin{figure}[pos=!htbp]
    \centering
    \includegraphics[width=\columnwidth]{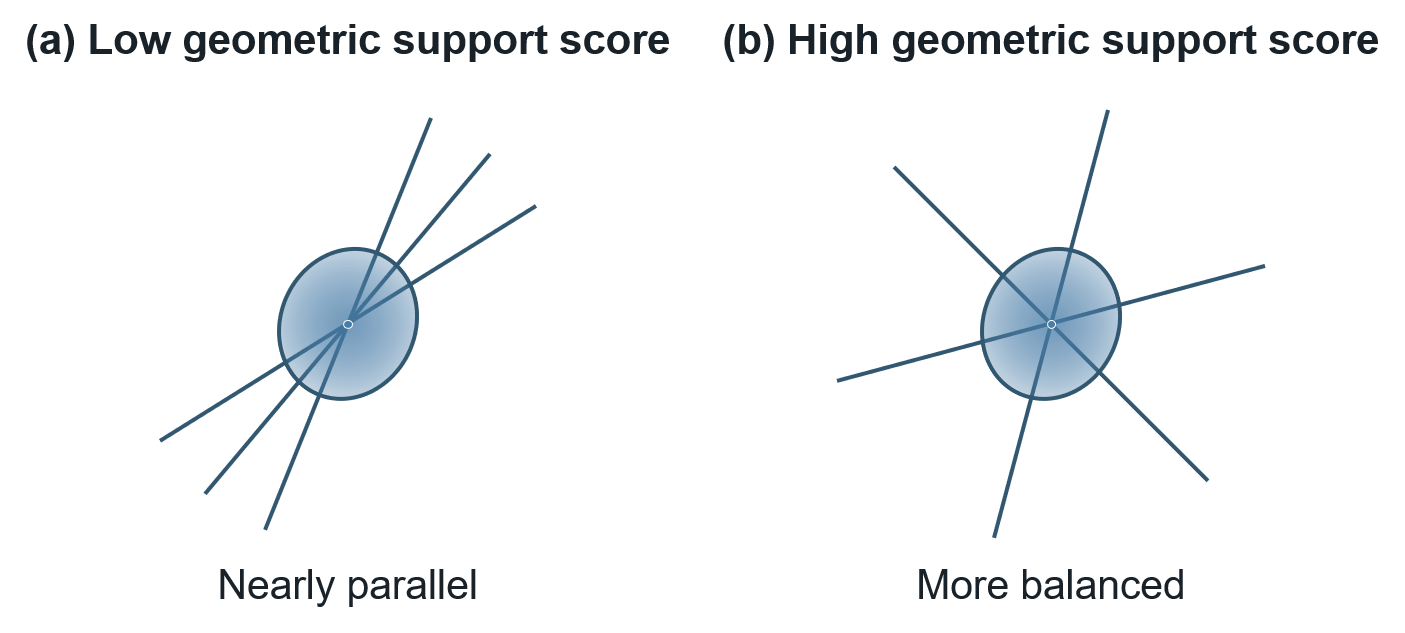}
    \caption{Geometric support score with different viewing-ray conditions. (a) Nearly parallel rays leave a common direction weakly constrained, yielding a small \(\lambda_{\min}(\mathbf F_n)\) and a low geometric support score \(s_n\). (b) More balanced directional constraints yield a larger \(\lambda_{\min}(\mathbf F_n)\) and a higher \(s_n\).}
    \label{fig:support}
\end{figure}

Since \(\mathbf F_n\) is symmetric and positive semidefinite, its eigenvalues characterize support along three principal directions. By the Rayleigh–Ritz characterization \citep{Horn_Johnson_2012}, the smallest eigenvalue equals the minimum quadratic value over all unit directions, and its associated eigenvector identifies the weakest supported direction. We define the geometric support score as:
\begin{equation}
    s_n
=
\frac{
\lambda_{\min}(\mathbf{F}_n)
}{
\operatorname{tr}(\mathbf{F}_n)+\epsilon
}
\end{equation}
where \(\epsilon>0\) ensures numerical stability. Trace normalization reduces sensitivity to the overall tensor magnitude and emphasizes the balance of directional constraints. As illustrated in Fig. \ref{fig:support}, this score is small when all viewing rays are nearly parallel (yielding a narrow angular range), and increases when support becomes more balanced across various directions.

During optimization, Gaussian positions evolve dynamically and viewing rays are processed sequentially. To retain historical observation information while adapting to the evolving geometry, we update equation \ref{eq:w} using an exponential moving average (EMA), analogous to the accumulation of historical moments in Adam optimizer \citep{kingma2014adam}:
\begin{equation}
\mathbf{F}_n^{(k)}
=
\gamma\mathbf{F}_n^{(k-1)}
+
(1-\gamma)\mathbf{A}_{n j_k}^{(k)},
\end{equation}
where \(j_k\) denotes the sampled camera at iteration $k$ and \(0\leq\gamma<1\) controls the retention of historical observations.

Upon the optimization is finished, the final support scores are computed for all Gaussians from their accumulated tensors. Gaussians with scores below a threshold \(\tau_s\) are automatically pruned. This criterion enables automatic removal of weakly constrained Gaussians, particularly in the peripheral regions where severely limited viewing-ray diversity leads to reconstruction failures apparent in orthogonal views.

\subsection{Optimizer-State Inheritance}
\label{sec:3.3}
The depth ambiguity analyzed in Sec. \ref{sec:3.2} leaves Gaussian positions weakly constrained, creating a vulnerable condition for geometric drift and floaters rather than directly causing them. The direct catalyst for these artifacts lies in abnormally amplified parameter updates. We identify an inherent flaw in how vanilla 3DGS initializes Adam optimizer states \citep{kingma2014adam} during densification as the underlying mechanism that amplifies these updates.

For a scalar Gaussian parameter \(\theta\), the Adam optimizer maintains biased first-moment estimate $m_t$ and biased second raw-moment estimate $v_t$ of its gradient \(g_t\):
\begin{equation}
    \begin{aligned}
m_t &= \beta_1 m_{t-1} + (1-\beta_1)g_t \\
v_t &= \beta_2 v_{t-1} + (1-\beta_2)g_t^2
\end{aligned}
\end{equation}
where $\beta_1$ and $\beta_2$ are the exponential decay rates. Starting from zero-initialized moments, bias correction is applied to compensate for their initial underestimation towards zero:
\begin{equation}
\begin{aligned}
\hat m_t &= \frac{m_t}{1-\beta_1^t}\\
\hat v_t &= \frac{v_t}{1-\beta_2^t}
\end{aligned}
\end{equation}
The parameter update is then computed as:

\begin{equation}
    \Delta\theta_t
=
-\eta_t\frac{\hat m_t}{\sqrt{\hat v_t}+\epsilon}
\end{equation}

where \(\eta_t\) is the learning rate and \(\epsilon>0\) is a small constant ensuring numerical stability.

In the densification process of vanilla 3DGS, newly created Gaussians are appended to existing parameter tensors. Their moment estimates \((m_t,v_t)\) are initialized to zero, whereas the shared optimizer step count $t$ is retained. Consequently, the bias correction mechanism uses the accumulated step count of the existing optimization process, despite the absence of a corresponding moment history for the new Gaussians. We term this phenomenon \textit{moment-step misalignment}.

\begin{figure*}[htbp]
    \centering
    \includegraphics[width=\textwidth]{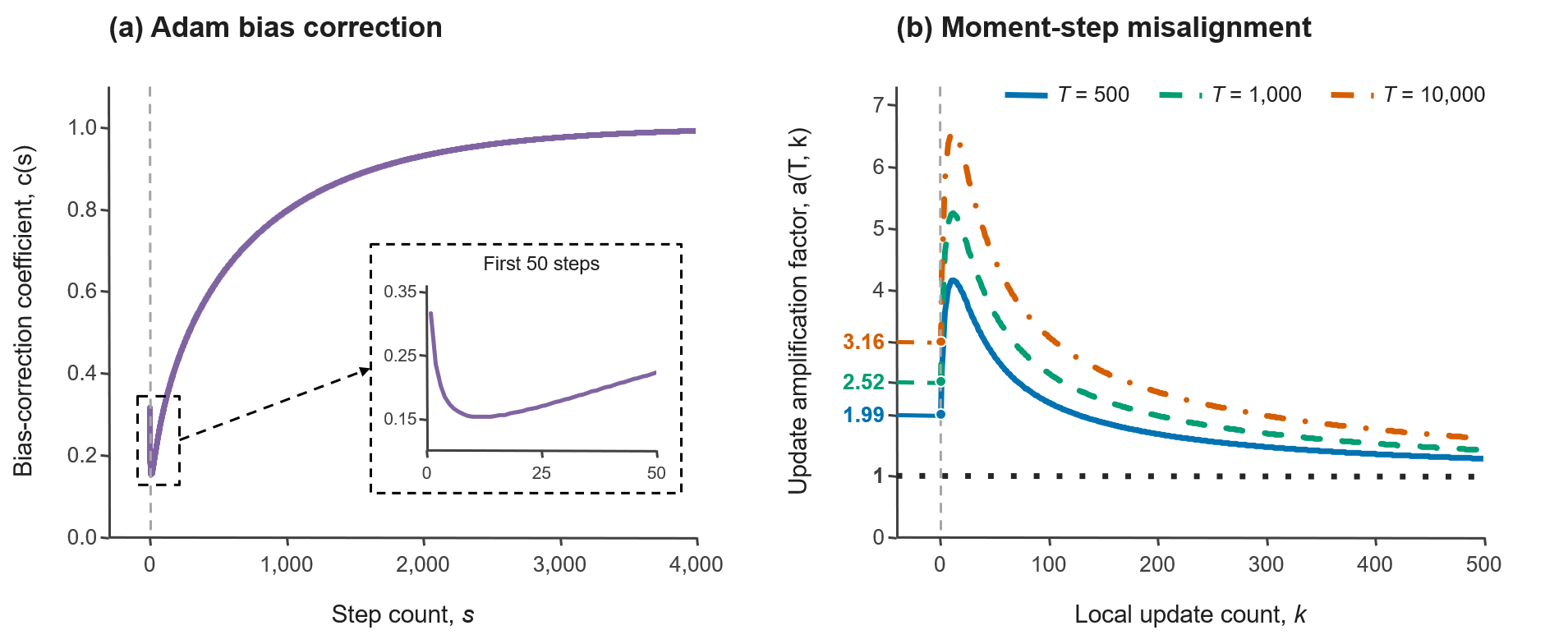}
    \caption{Parameter-update amplification caused by \textit{moment-step misalignment} in Adam \citep{kingma2014adam}. (a) Bias-correction coefficient \(c(s)\) as a function of the step count \(s\), with an inset showing the first 50 steps. (b) Relative parameter-update scaling factor \(a(T,k)=c(T+k)/c(k)\), where \(T\) is the optimizer step count at Gaussian creation and \(k\) is the number of subsequent updates.}
    \label{fig:adam}
\end{figure*}

To examine its effect, let \(T\) denote the step count at the instant a Gaussian is created, and \(k\geq1\) be the number of subsequent updates received by that Gaussian. Its accumulated global step count is therefore \(t=T+k\), whereas the number of updates it has actually experienced since creation is only \(k\). Neglecting \(\epsilon\) when it is small relative to the denominator, we define the bias-correction coefficient as a function of the step count $s$, also shown in Fig. \ref{fig:adam} (a):

\begin{equation}
    c(s)=\frac{\sqrt{1-\beta_2^s}}{1-\beta_1^s}
\end{equation}

Since Adam’s bias correction strictly requires the step count to match the moment history, we compare the original (misaligned) and moment-step-aligned reference updates using identical moments \(m_k,v_k\) and learning rate \(\eta_{T+k}\): 
 
\begin{equation}
\begin{aligned}
\Delta\theta_{T,k}^{\mathrm{original}}
&\approx -\eta_{T+k}\frac{m_k}{\sqrt{v_k}}\cdot c(T+k),\\
\Delta\theta_{T,k}^{\mathrm{aligned}}
&\approx -\eta_{T+k}\frac{m_k}{\sqrt{v_k}}\cdot c(k).
\end{aligned}
\end{equation}

We define the relative parameter-update scaling factor as:
\begin{equation}
    a(T,k)
=
\frac{c(T+k)}{c(k)}
=
\frac{1-\beta_1^k}{1-\beta_1^{T+k}}
\sqrt{\frac{1-\beta_2^{T+k}}{1-\beta_2^k}}
\end{equation}

This relative scaling directly quantifies the degree of parameter-update amplification caused by \textit{moment-step misalignment}. Using Adam’s default decay rates, \(\beta_1=0.9\) and \(\beta_2=0.999\), we illustrate this amplification in Fig. \ref{fig:adam} (b). Compared with aligned bias correction, the original updates can be amplified by up to seven times, and this abnormal amplification persists over hundreds of subsequent updates.

In scenes captured from surrounding viewpoints with rich viewing diversity, strong geometric constraints can effectively restrict the displacement caused by amplified updates, preventing them from developing into geometric drift and floaters. However, such artifacts arise much more readily in large-scale UAV datasets due to two compounding factors. First, UAV images are predominantly acquired from oblique or nadir views, often yielding poorly conditioned viewing ray geometries (as analyzed in Sec. \ref{sec:3.2}). Second, under the synchronous optimization scheme \citep{ding2026step} of vanilla 3DGS, a Gaussian invisible in the current training view can still undergo parameter updates driven by its accumulated moment estimates \citep{ding2026step}. Because UAV datasets cover extensive areas, an individual Gaussian is typically observed in only a small subset of the training images. During a single pass through all training views, it may therefore undergo multiple updates without photometric supervision. These repeated, abnormally amplified updates can accumulate displacement along weakly constrained directions, severely exacerbating geometric drift and floaters.

To mitigate this parameter-update amplification, two moment-step alignment strategies can be considered: initializing each child Gaussian with zero moments and resetting the step count to zero, or inheriting the parent Gaussian’s accumulated moment estimates while retaining the global step count. We adopt the latter to preserve the optimization trajectory and moment information of the parent. As illustrated in Fig. \(\ref{fig:framework}\)(B), a child Gaussian \(c\) inherits the accumulated moment estimates of its parent \(p\) at the moment of creation:
\begin{equation}
    m_c^{\mathrm{birth}}\leftarrow m_p
\qquad
v_c^{\mathrm{birth}}\leftarrow v_p
\end{equation}

while the shared optimizer step count \(T\) is retained. By maintaining strict moment-step alignment, this strategy mitigates abnormal amplification of Gaussian parameter updates, thereby suppressing geometric drift and floaters without introducing any external priors into the training process.

\subsection{Robust Gravity Direction Estimation}
\label{sec:3.4}
The high-fidelity scene geometry of a generated TDOM must be paired with a reliable orthogonal projection reference. Existing 3DGS-based approaches either require manual rotational correction or rely on camera-distribution priors, which may yield unreliable orthogonal projection references when complex datasets consist exclusively of oblique views or exhibit substantial variations in flight altitude. To address these limitations, we develop a robust gravity direction estimation solution that integrates image-derived gravity cues, camera-distribution priors, and explicit nadir views through confidence-based selection and fusion mechanism, as illustrated in Fig. \ref{fig:framework} (C).

\subsubsection{Image-Derived Gravity Direction Estimation}
\label{3.4.1}
Perspective Fields \citep{jin2023perspective} recovers camera calibration parameters from a single image using pretrained neural networks. It predicts dense up-vector and latitude fields to represent local image perspective, and maps these fields to camera parameters through a dedicated network - ParamNet. We employ this model to extract image-derived gravity cues from individual UAV images. For an input image \(I_j\), the orientation prediction is expressed as

\begin{equation}
\mathcal F:\ I_j\mapsto(\phi_j,\theta_j)
\end{equation}

where \(\mathcal F\) denotes the pretrained network. The roll angle \(\phi_j\) describes rotation about the camera optical axis, while the pitch angle \(\theta_j\) specifies the elevation of the optical axis relative to the horizontal plane.

The predicted orientation is combined with the SfM-derived camera rotation to obtain a gravity direction (down-vector) in the world coordinate system:
\begin{equation}
    \mathbf d_j
=
\mathbf W_j^\top
\mathbf R_z(-\phi_j)
\mathbf R_x\!\left(-\theta_j-\frac{\pi}{2}\right)
\mathbf e_z,
\qquad
\mathbf e_z=[0,0,1]^\top
\end{equation}

where \(\mathbf W_j\) is the world-to-camera rotation matrix, and \(\mathbf e_z\) is the optical-axis vector in the camera coordinate system, and the matrices \(\mathbf R_z\) and \(\mathbf R_x\) encode the roll and pitch corrections, respectively. 

However, independent single-image predictions do not enforce cross-view consistency, often resulting in scattered gravity direction estimates. To address this, as illustrated in Fig. \ref{fig:framework} (C-Robust consensus), we employ a robust statistical filtering strategy based on directional statistics. An initial reference is selected as the predicted direction with the smallest median angular distance to all other predictions. Directions with angular deviations exceeding an adaptive threshold (computed using the median and the Median Absolute Deviation \citep{rousseeuw1993alternatives}) are rejected as gross outliers. The remaining inlier directions are averaged to obtain a refined reference, around which a second screening using Median Absolute Deviation and an upper-quantile threshold removes further inconsistent predictions.

Let \(\mathcal I\) denote the resulting inlier set. The final image-derived gravity estimate is computed as:
\begin{equation}
    \mathbf d_{\mathrm I}
=
\frac{\sum_{j\in\mathcal I}\mathbf d_j}
{\left\|\sum_{j\in\mathcal I}\mathbf d_j\right\|}
\end{equation}
Its confidence is evaluated using the inlier count, inlier ratio, angular deviations, and directional concentration:
\begin{equation}
    r_{\mathrm I}=\frac{|\mathcal I|}{M},
\qquad
e_j=\arccos\!\left(\mathbf d_j^\top\mathbf d_{\mathrm I}\right),
\qquad
\ell_{\mathrm I}
=
\left\|
\frac{1}{|\mathcal I|}
\sum_{j\in\mathcal I}\mathbf d_j
\right\|
\end{equation}
where \(M\) is the number of input images and \(\ell_{\mathrm I}\) is the mean resultant length from directional statistics. A value of \(\ell_{\mathrm I}\) close to one indicates strong directional agreement among the inliers. The estimate is classified as \textit{\textbf{valid}} only when all confidence metrics satisfy the prescribed criteria, ensuring that only highly consistent image-derived estimates participate in the subsequent fusion.

\subsubsection{Camera-Distribution-Based Gravity Direction Estimation}
\label{sec:3.4.2}
Under the prior assumption of constant flight altitude, the normal to a plane fitted to the camera centers serves as a candidate gravity direction. We employ RANSAC \citep{fischler1981random} to identify camera centers consistent with a dominant plane, and apply principal component analysis (PCA) to the retained inliers to extract its normal vector \(\mathbf d_{\mathrm P}\).

The confidence of this estimate is evaluated based on the plane-fitting quality and the spatial distribution of the supporting cameras. Specifically, we consider the inlier ratio \(r_{\mathrm P}\), the median and 90th-percentile point-to-plane residuals. Furthermore, we analyze the covariance structure of the centered inlier coordinates via Singular Value Decomposition (SVD), yielding singular values $\sigma_1 \geq \sigma_2 \geq \sigma_3$. We define two critical spatial ratios:
\begin{equation}
q_{\mathrm{spread}}=\frac{\sigma_2}{\sigma_1}
\qquad
q_{\mathrm{thickness}}=\frac{\sigma_3}{\sigma_2}
\end{equation}
A small \(q_{\mathrm{spread}}\) indicates a nearly collinear camera distribution, for which the plane normal is poorly constrained. Conversely, a large \(q_{\mathrm{thickness}}\) indicates substantial departure from a planar distribution (i.e., substantial altitude variations), violating the constant-altitude prior. The camera-distribution estimate is classified as \textit{\textbf{valid}} when these metrics confirm sufficient planarity and spatial spread.

\subsubsection{Confidence-Based Selection and Fusion}
\label{sec:3.4.3}
\begin{figure*}[htbp]
    \centering
    \includegraphics[width=\textwidth]{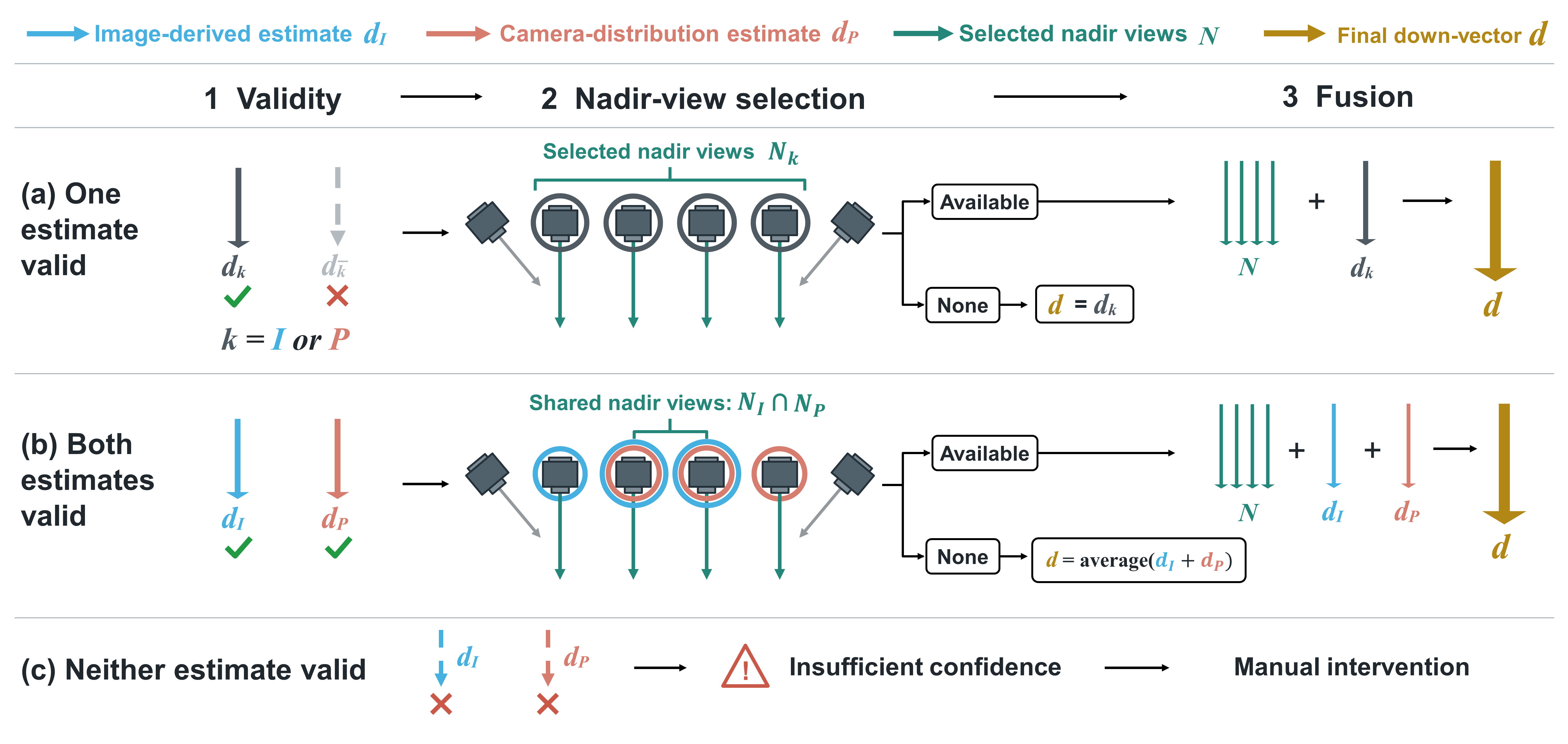}
    \caption{Confidence-based selection and fusion for robust gravity direction estimation.}
    \label{fig:gravity}
\end{figure*}

As illustrated in Fig. \ref{fig:gravity}, we formulate the final gravity estimation as a confidence-driven multi-source fusion problem. Let $\mathbf{d}_{\mathrm{I}}$ and $\mathbf{d}_{\mathrm{P}}$ be the image-derived and camera-distribution-based estimates, respectively. Additionally, we leverage explicit nadir UAV cameras as highly reliable downward references, as their optical axes are inherently aligned with the gravity vector. A camera is classified as a nadir view if its optical axis aligns closely with a valid gravity estimate (e.g., the intersection angle is below a predefined threshold). Let $\mathcal{N}_{\mathrm{I}}$ and $\mathcal{N}_{\mathrm{P}}$ denote the sets of nadir views identified by $\mathbf{d}_{\mathrm{I}}$ and $\mathbf{d}_{\mathrm{P}}$, respectively.

The fusion strategy adaptively combines these sources based on their \textit{validity} status:
\begin{itemize}
    \item \textbf{Only one estimate is valid:} The valid estimate $\mathbf{d}_{\mathrm{k}}$ ($k=\mathrm{I}\text{ or }\mathrm{P}$) is averaged with the optical axes of its identified nadir views ($\mathcal{N}_{\mathrm{k}}$). If no nadir views are available, it is used directly.
    \item \textbf{Both estimates are valid:} We compute the intersection of their corresponding nadir view sets, $\mathcal{N}_{\mathrm{I}} \cap \mathcal{N}_{\mathrm{P}}$. If the intersection is non-empty, the final down-vector is obtained by averaging $\mathbf{d}_{\mathrm{I}}$, $\mathbf{d}_{\mathrm{P}}$, and the optical axes of the shared nadir views, providing a highly robust consensus. If the intersection is empty, $\mathbf{d}_{\mathrm{I}}$ and $\mathbf{d}_{\mathrm{P}}$ are directly averaged.
    \item \textbf{Neither estimate is valid:} The system issues a warning, indicating that the dataset lacks sufficient constraints for automatic gravity estimation, thereby prompting manual intervention.
\end{itemize}

The resulting consensus gravity-aligned down-vector establishes a highly reliable orthogonal projection reference for rendering the pruned 3D Gaussian field into a TDOM.

\section{Experiments}
\label{sec:experiments}
To evaluate the efficacy of the proposed Tortho-Gaussian++ in generating TDOMs, the reported experiments comprise five parts. Qualitative comparisons examine geometric drift, floaters, and tilt-induced facade artifacts (Sec. \ref{sec:qualitative}), while quantitative evaluations assess rendering quality, time efficiency, resource consumption and planimetric accuracy (Sec. \ref{sec:quantitative}). We further investigate the robustness of gravity direction estimation under challenging acquisition conditions (Sec. \ref{sec:robustness}) and conduct ablation studies to assess the individual contributions of the proposed components (Sec. \ref{sec:ablation}). Finally, we demonstrate the integration of our optimizer-state inheritance strategy into existing Gaussian splatting pipelines and evaluate its benefits (Sec. \ref{sec:integration}).

\begin{figure*}[!t]
    \centering
    \includegraphics[width=\textwidth]{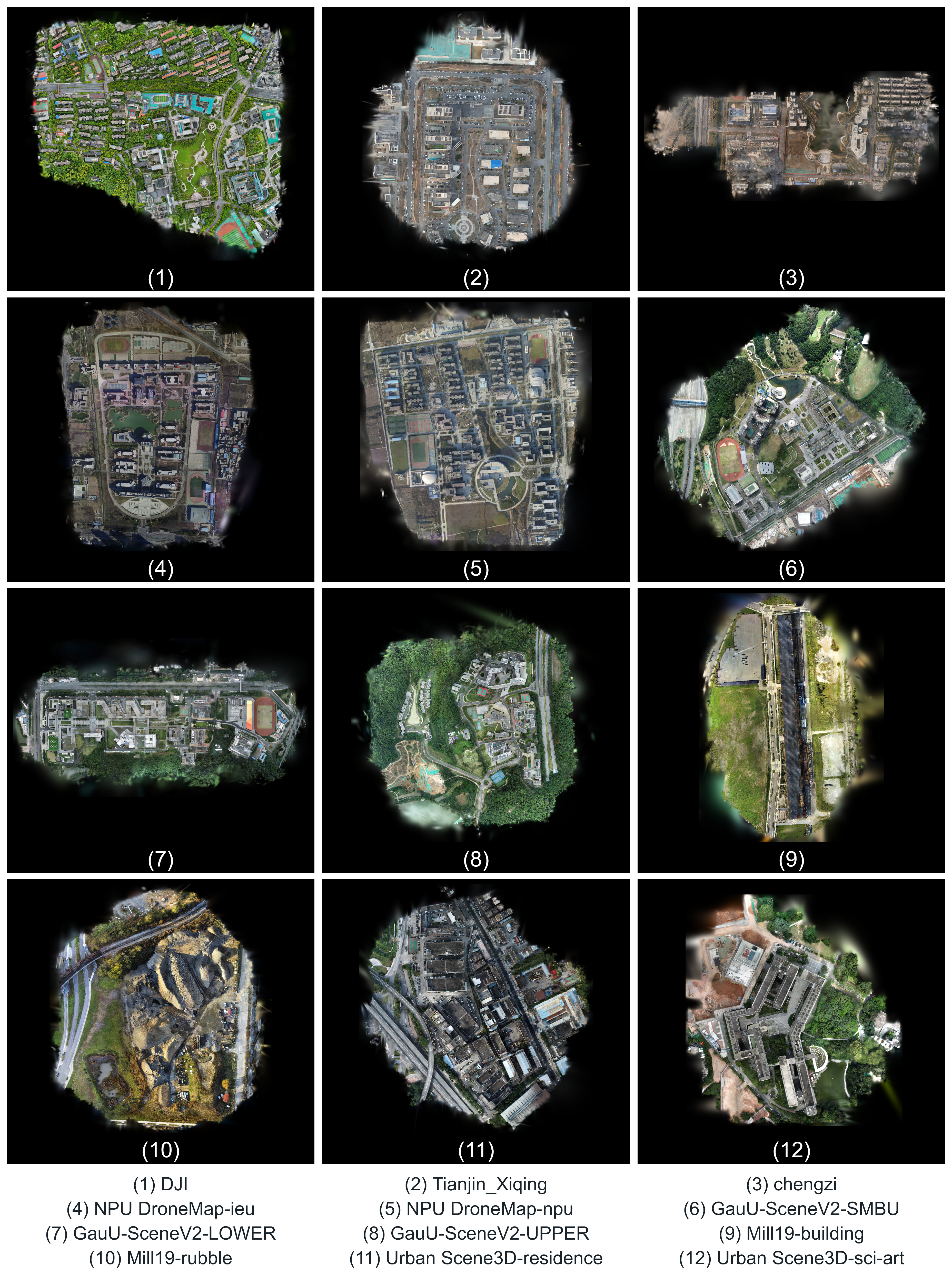}
    \caption{The complete TDOMs generated by the proposed Tortho-Gaussian++ across diverse large-scale UAV scenes.}
    \label{fig:Overall}
\end{figure*}

\begin{table*}[!t]
\centering
\begin{threeparttable}

\captionsetup{
    format=plain,
    justification=raggedright,
    singlelinecheck=false,
    labelsep=newline,
    margin=0pt
}

\caption{Details of the experimental datasets.}
\label{tab:datasets}

\small
\setlength{\tabcolsep}{5pt}
\renewcommand{\arraystretch}{1.15}

\begin{tabular}{@{}llcccc@{}}
\toprule
Dataset & Scene & Images & Resolution
& Viewing configuration & Flight altitude \\
\midrule

\multirow{4}{*}{Custom-collected}
& Chengzi    & 997 & $8192 \times 5468$
& Oblique & Stable \\
& DJI        & 874 & $4032 \times 3024$
& Nadir & Stable \\
& Tianjin\_Xiqing & 305 & $4000 \times 3000$
& Nadir + oblique & Stable \\
& Hanyang    & 54  & $5280 \times 3956$
& Nadir + oblique & Stable \\

\midrule
\multirow{2}{*}{NPU DroneMap}
& ieu & 467 & $1920 \times 1080$
& Nadir & Stable \\
& npu & 457 & $1920 \times 1080$
& Nadir & Stable \\

\midrule
\multirow{3}{*}{GauU-Scene V2}
& CUHK\_LOWER & 670 & $5474 \times 3643$
& Nadir + oblique & Stable \\
& CUHK\_UPPER & 715 & $5469 \times 3638$
& Nadir + oblique & Stable \\
& SMBU        & 563 & $5468 \times 3636$
& Nadir + oblique & Stable \\

\midrule
\multirow{2}{*}{Mill 19}
& Building & 1,940 & $4608 \times 3456$
& Oblique & Stable \\
& Rubble   & 1,678 & $4608 \times 3456$
& Oblique & Stable \\

\midrule
\multirow{2}{*}{UrbanScene3D}
& Residence & 2,582 & $5472 \times 3648$
& Oblique & Variable \\
& Sci-Art   & 3,019 & $4864 \times 3648$
& Oblique & Variable \\

\bottomrule
\end{tabular}

\end{threeparttable}
\end{table*}

\subsection{Experimental Settings}
\label{sec:settings}
\textbf{Datasets.} We conduct extensive experiments on multiple large-scale UAV datasets, including custom-collected scenes and selected scenes from four public benchmarks: NPU DroneMap \citep{bu2016map2dfusion}, GauU-Scene V2 \citep{xiong2024gauu}, Mill 19 \citep{turki2022mega}, and UrbanScene3D \citep{lin2022capturing}. The overall TDOMs generated on these tested datasets are shown in Fig. \ref{fig:Overall}, and detailed acquisition configurations are summarized in Table \ref{tab:datasets}. These datasets encompass diverse buildings and other man-made structures, enabling a comprehensive evaluation of geometric drift and floater suppression. To assess the robustness of gravity direction estimation, the datasets cover diverse acquisition conditions, including nadir-only, mixed nadir and oblique, and oblique-only imagery, as well as substantial variations in flight altitude in some scenes. In particular, the \textit{Hanyang} scene contains high-rise buildings, which place more stringent demands on the accuracy of the gravity-aligned orthogonal projection reference. Given its limited spatial extent, this scene is used only for the qualitative evaluation of tilt-induced facade artifacts.

\textbf{Evaluation Criteria.} We evaluate novel view synthesis quality using three widely used image quality metrics: peak signal-to-noise ratio (PSNR), structural similarity index measure (SSIM) \citep{wang2004image}, and learned perceptual image patch similarity (LPIPS) \citep{zhang2018unreasonable}. We also report training time, peak GPU memory usage (VRAM), and the final Gaussian number to assess computational efficiency, memory consumption, and model compactness. In addition, the \textit{npu} scene provides ground control points, enabling evaluation of the planimetric accuracy of the generated TDOMs using the root mean square error (RMSE).

\textbf{Comparative Methods.} We compare Tortho-Gaussian++ with two groups of methods: general-purpose Gaussian Splatting methods, including 3DGS \citep{kerbl20233d}, CityGaussian \citep{liu2024citygaussian}, and FastGS \citep{ren2026fastgs}; and TDOM-oriented methods, including Tortho-Gaussian \citep{wang2026tortho} and Ortho-3DGS \citep{yang2025ortho}. Additionally, we compare our method with traditional photogrammetric methods in planimetric accuracy, including Map2DFusion \citep{bu2016map2dfusion}, ContextCapture \citep{bentley2022contextcapture}, Metashape \citep{agisoft2022metashape} and Pix4DMapper \citep{pix4d2022mapper}.

\textbf{Implementation Details.} For each scene, SfM is performed on all images to obtain camera poses and a sparse point cloud. The images are then divided into training and testing sets for Gaussian optimization. All scenes are trained for 30,000 iterations. The EMA decay factor for geometric support accumulation is set to $\gamma=0.95$. The automatic pruning of unreliable peripheral regions employs an empirical threshold $\tau_s$ of 0.1. The validity criteria for robust gravity direction estimation are specified as follows. The image-derived estimate $\mathbf d_{\mathrm I}$ must satisfy $r_{\mathrm I}\geq 0.50$ and $\ell_{\mathrm I}\geq 0.97$, with the median and 90th percentile of the inlier angular deviations $e_j$ not exceeding $10^\circ$ and $16^\circ$, respectively. The camera-distribution-based estimate $\mathbf d_{\mathrm P}$ must satisfy $r_{\mathrm P}\geq 0.60$, $q_{\mathrm{spread}}\geq 0.05$, and $q_{\mathrm{thickness}}\leq 0.15$, ensuring sufficient inlier support, spatial spread, and planarity.

For qualitative evaluation of geometric drift and floaters, the Gaussian scenes reconstructed by all methods are orthogonally rendered using our projection reference to ensure consistent projection conditions. For qualitative evaluation of tilt-induced facade artifacts, we compare only the TDOM-oriented methods on datasets where all compared methods can establish a reasonable projection reference, with each method using its own rendering procedure. For quantitative evaluation, all comprehensive metrics of Tortho-Gaussian++ are recorded before automatic pruning. For quantitative evaluation of planimetric accuracy, the general-purpose methods render TDOMs with our projection reference and the TDOM-oriented methods use their own rendering procedures. All experiments are conducted on an NVIDIA RTX 4090 GPU with 48 GB of VRAM.

\definecolor{qbestcolor}{HTML}{FFA99B}
\definecolor{qsecondcolor}{HTML}{FFD39B}
\definecolor{qthirdcolor}{HTML}{FFF5AD}
\providecommand{\qbest}[1]{\cellcolor{qbestcolor}\textbf{#1}}
\providecommand{\qsecond}[1]{\cellcolor{qsecondcolor}#1}
\providecommand{\qthird}[1]{\cellcolor{qthirdcolor}#1}

\begin{table*}[!t]
\centering
\captionsetup{format=plain,justification=raggedright,
  singlelinecheck=false,labelsep=newline,margin=0pt}
\caption{Quantitative comparison of novel view synthesis quality, time efficiency and resource consumption on twelve UAV scenes. The best, second-best, and third-best results are highlighted in red, orange, and yellow, respectively. Rankings are determined before rounding.}
\label{tab:quantitative}
\begingroup
\scriptsize
\setlength{\tabcolsep}{2pt}
\renewcommand{\arraystretch}{1.12}
\resizebox{\textwidth}{!}{%
\begin{tabular}{@{}l|cccccc|cccccc|cccccc@{}}
\toprule
\multicolumn{1}{@{}c|}{\multirow{2}{*}{Method}} & \multicolumn{6}{c|}{Chengzi} & \multicolumn{6}{c|}{DJI} & \multicolumn{6}{c}{Tianjin\_Xiqing} \\
\cmidrule(lr){2-7}\cmidrule(lr){8-13}\cmidrule(l){14-19}
 & PSNR$\uparrow$ & SSIM$\uparrow$ & LPIPS$\downarrow$ & Time$\downarrow$ & $N_{\mathrm{GS}}\downarrow$ & VRAM$\downarrow$ & PSNR$\uparrow$ & SSIM$\uparrow$ & LPIPS$\downarrow$ & Time$\downarrow$ & $N_{\mathrm{GS}}\downarrow$ & VRAM$\downarrow$ & PSNR$\uparrow$ & SSIM$\uparrow$ & LPIPS$\downarrow$ & Time$\downarrow$ & $N_{\mathrm{GS}}\downarrow$ & VRAM$\downarrow$ \\
\midrule
3DGS & 25.04 & 0.874 & 0.157 & \qthird{53.00} & \qthird{14.63M} & 23.03\,GiB & \qthird{24.54} & \qthird{0.863} & \qsecond{0.104} & \qthird{62.38} & \qthird{17.89M} & \qthird{29.07\,GiB} & \qthird{26.56} & \qbest{0.866} & \qthird{0.124} & \qthird{30.08} & 4.56M & 7.29\,GiB \\
CityGaussian & 26.01 & \qthird{0.893} & \qbest{0.119} & 342.93 & 19.91M & 28.19\,GiB & \multicolumn{6}{c|}{\textit{OOM}} & 26.52 & \qthird{0.865} & \qbest{0.118} & 309.48 & 5.01M & 8.87\,GiB \\
FastGS & \qsecond{28.39} & 0.888 & 0.184 & \qbest{4.28} & \qbest{1.15M} & \qbest{2.15\,GiB} & \qsecond{26.84} & \qsecond{0.891} & \qthird{0.117} & \qbest{3.23} & \qbest{0.56M} & \qbest{0.99\,GiB} & 26.45 & 0.852 & 0.173 & \qbest{6.75} & \qbest{0.53M} & \qbest{1.10\,GiB} \\
Tortho-Gaussian & 25.20 & 0.875 & \qthird{0.156} & 79.35 & 19.20M & \qthird{17.03\,GiB} & \multicolumn{6}{c|}{\textit{OOM}} & 24.58 & 0.823 & 0.199 & 40.60 & \qthird{1.42M} & \qthird{3.94\,GiB} \\
Ortho-3DGS & \qthird{27.85} & \qsecond{0.895} & \qsecond{0.143} & 71.63 & 16.72M & 32.22\,GiB & \multicolumn{6}{c|}{\textit{OOM}} & \qsecond{26.58} & \qsecond{0.866} & \qsecond{0.124} & 35.02 & 4.61M & 7.96\,GiB \\
\cmidrule(lr){1-19}
\textbf{Tortho-Gaussian++} & \qbest{29.01} & \qbest{0.901} & 0.158 & \qsecond{33.80} & \qsecond{7.15M} & \qsecond{11.40\,GiB} & \qbest{27.49} & \qbest{0.914} & \qbest{0.083} & \qsecond{27.67} & \qsecond{6.87M} & \qsecond{11.12\,GiB} & \qbest{26.74} & 0.860 & 0.160 & \qsecond{18.00} & \qsecond{1.15M} & \qsecond{3.31\,GiB} \\
\midrule
\multicolumn{1}{@{}c|}{\multirow{2}{*}{Method}} & \multicolumn{6}{c|}{ieu} & \multicolumn{6}{c|}{npu} & \multicolumn{6}{c}{CUHK\_LOWER} \\
\cmidrule(lr){2-7}\cmidrule(lr){8-13}\cmidrule(l){14-19}
 & PSNR$\uparrow$ & SSIM$\uparrow$ & LPIPS$\downarrow$ & Time$\downarrow$ & $N_{\mathrm{GS}}\downarrow$ & VRAM$\downarrow$ & PSNR$\uparrow$ & SSIM$\uparrow$ & LPIPS$\downarrow$ & Time$\downarrow$ & $N_{\mathrm{GS}}\downarrow$ & VRAM$\downarrow$ & PSNR$\uparrow$ & SSIM$\uparrow$ & LPIPS$\downarrow$ & Time$\downarrow$ & $N_{\mathrm{GS}}\downarrow$ & VRAM$\downarrow$ \\
\midrule
3DGS & 32.41 & \qthird{0.936} & \qthird{0.088} & \qthird{47.05} & 9.83M & 15.42\,GiB & \qthird{36.38} & \qthird{0.968} & \qthird{0.040} & \qthird{40.72} & 7.46M & 12.68\,GiB & 27.56 & 0.892 & \qthird{0.128} & \qthird{32.97} & \qthird{6.96M} & \qthird{10.64\,GiB} \\
CityGaussian & \multicolumn{6}{c|}{\textit{Failed}} & \multicolumn{6}{c|}{\textit{Failed}} & \qthird{28.08} & \qbest{0.900} & \qbest{0.107} & 207.92 & 8.32M & 12.72\,GiB \\
FastGS & 32.49 & 0.921 & 0.133 & \qbest{4.34} & \qbest{0.67M} & \qbest{1.25\,GiB} & 35.27 & 0.956 & 0.070 & \qbest{4.24} & \qbest{0.41M} & \qbest{0.84\,GiB} & 27.74 & 0.871 & 0.184 & \qbest{4.68} & \qbest{0.76M} & \qbest{1.51\,GiB} \\
Tortho-Gaussian & \qthird{32.69} & \qsecond{0.936} & \qsecond{0.088} & 48.94 & \qthird{4.84M} & \qthird{8.11\,GiB} & 36.27 & \qsecond{0.968} & \qsecond{0.040} & 43.18 & \qthird{4.99M} & \qthird{8.74\,GiB} & 27.47 & 0.891 & 0.129 & 78.80 & 11.40M & 27.66\,GiB \\
Ortho-3DGS & \qsecond{32.74} & \qbest{0.936} & \qbest{0.086} & 54.59 & 11.07M & 18.83\,GiB & \qsecond{36.44} & 0.967 & \qbest{0.040} & 42.91 & 8.84M & 15.35\,GiB & \qsecond{28.24} & \qthird{0.896} & \qsecond{0.120} & 49.96 & 8.69M & 13.39\,GiB \\
\cmidrule(lr){1-19}
\textbf{Tortho-Gaussian++} & \qbest{32.79} & 0.932 & 0.104 & \qsecond{20.28} & \qsecond{3.22M} & \qsecond{6.30\,GiB} & \qbest{36.50} & \qbest{0.969} & 0.045 & \qsecond{21.62} & \qsecond{2.82M} & \qsecond{5.88\,GiB} & \qbest{28.40} & \qsecond{0.899} & 0.140 & \qsecond{16.25} & \qsecond{2.63M} & \qsecond{6.07\,GiB} \\
\midrule
\multicolumn{1}{@{}c|}{\multirow{2}{*}{Method}} & \multicolumn{6}{c|}{CUHK\_UPPER} & \multicolumn{6}{c|}{SMBU} & \multicolumn{6}{c}{Building} \\
\cmidrule(lr){2-7}\cmidrule(lr){8-13}\cmidrule(l){14-19}
 & PSNR$\uparrow$ & SSIM$\uparrow$ & LPIPS$\downarrow$ & Time$\downarrow$ & $N_{\mathrm{GS}}\downarrow$ & VRAM$\downarrow$ & PSNR$\uparrow$ & SSIM$\uparrow$ & LPIPS$\downarrow$ & Time$\downarrow$ & $N_{\mathrm{GS}}\downarrow$ & VRAM$\downarrow$ & PSNR$\uparrow$ & SSIM$\uparrow$ & LPIPS$\downarrow$ & Time$\downarrow$ & $N_{\mathrm{GS}}\downarrow$ & VRAM$\downarrow$ \\
\midrule
3DGS & 26.82 & 0.886 & \qthird{0.114} & 49.78 & \qthird{7.37M} & \qthird{11.07\,GiB} & \qthird{25.55} & \qthird{0.893} & 0.108 & \qthird{32.08} & 5.48M & \qthird{8.56\,GiB} & 22.84 & \qthird{0.821} & \qsecond{0.170} & \qthird{44.97} & \qthird{7.92M} & \qthird{11.88\,GiB} \\
CityGaussian & \qthird{27.20} & \qbest{0.896} & \qbest{0.098} & 214.05 & 9.33M & 14.41\,GiB & 25.44 & \qsecond{0.895} & \qbest{0.100} & 173.10 & 5.95M & 9.51\,GiB & \qsecond{23.44} & \qbest{0.846} & \qbest{0.145} & 357.02 & 9.24M & 14.79\,GiB \\
FastGS & 26.75 & 0.858 & 0.179 & \qbest{4.95} & \qbest{0.85M} & \qbest{1.46\,GiB} & 25.26 & 0.867 & 0.163 & \qbest{5.03} & \qbest{0.67M} & \qbest{1.27\,GiB} & 22.88 & 0.798 & 0.223 & \qsecond{21.37} & \qbest{0.74M} & \qbest{2.55\,GiB} \\
Tortho-Gaussian & 26.79 & 0.885 & 0.115 & 47.51 & 8.68M & 14.61\,GiB & 25.49 & 0.883 & \qthird{0.107} & 46.52 & \qthird{4.97M} & 10.22\,GiB & 22.77 & 0.820 & 0.178 & 86.94 & 11.06M & 17.69\,GiB \\
Ortho-3DGS & \qbest{27.91} & \qsecond{0.891} & \qsecond{0.110} & \qthird{44.78} & 8.49M & 14.57\,GiB & \qbest{26.76} & \qbest{0.898} & \qsecond{0.104} & 35.62 & 5.95M & 9.24\,GiB & \qthird{23.35} & 0.810 & 0.179 & 71.43 & 9.78M & 14.61\,GiB \\
\cmidrule(lr){1-19}
\textbf{Tortho-Gaussian++} & \qsecond{27.60} & \qthird{0.886} & 0.136 & \qsecond{15.35} & \qsecond{2.35M} & \qsecond{5.36\,GiB} & \qsecond{26.24} & 0.890 & 0.127 & \qsecond{14.65} & \qsecond{1.84M} & \qsecond{4.58\,GiB} & \qbest{23.90} & \qsecond{0.826} & \qthird{0.170} & \qbest{20.64} & \qsecond{3.70M} & \qsecond{7.65\,GiB} \\
\midrule
\multicolumn{1}{@{}c|}{\multirow{2}{*}{Method}} & \multicolumn{6}{c|}{Rubble} & \multicolumn{6}{c|}{Residence} & \multicolumn{6}{c}{Sci-Art} \\
\cmidrule(lr){2-7}\cmidrule(lr){8-13}\cmidrule(l){14-19}
 & PSNR$\uparrow$ & SSIM$\uparrow$ & LPIPS$\downarrow$ & Time$\downarrow$ & $N_{\mathrm{GS}}\downarrow$ & VRAM$\downarrow$ & PSNR$\uparrow$ & SSIM$\uparrow$ & LPIPS$\downarrow$ & Time$\downarrow$ & $N_{\mathrm{GS}}\downarrow$ & VRAM$\downarrow$ & PSNR$\uparrow$ & SSIM$\uparrow$ & LPIPS$\downarrow$ & Time$\downarrow$ & $N_{\mathrm{GS}}\downarrow$ & VRAM$\downarrow$ \\
\midrule
3DGS & 26.87 & 0.870 & 0.161 & \qsecond{28.38} & \qthird{5.86M} & \qthird{9.00\,GiB} & 22.43 & 0.852 & \qthird{0.152} & \qsecond{29.65} & \qthird{5.69M} & \qthird{8.98\,GiB} & 22.16 & \qthird{0.867} & \qthird{0.144} & \qsecond{19.87} & 3.07M & \qthird{5.24\,GiB} \\
CityGaussian & \qbest{27.52} & \qbest{0.888} & \qbest{0.137} & 308.38 & 6.35M & 11.98\,GiB & \qbest{22.69} & \qbest{0.861} & \qbest{0.142} & 362.55 & 6.01M & 10.94\,GiB & \qsecond{22.36} & \qbest{0.875} & \qbest{0.135} & 255.63 & 2.95M & 5.41\,GiB \\
FastGS & 25.33 & 0.812 & 0.247 & \qbest{20.22} & \qbest{0.75M} & \qbest{2.62\,GiB} & 22.06 & 0.825 & 0.195 & \qthird{34.62} & \qbest{0.77M} & \qbest{3.05\,GiB} & 21.55 & 0.837 & 0.191 & \qthird{41.10} & \qbest{0.57M} & \qbest{3.54\,GiB} \\
Tortho-Gaussian & 26.65 & 0.849 & 0.196 & 59.95 & 6.59M & 9.48\,GiB & 22.49 & \qthird{0.852} & 0.152 & 83.16 & 6.56M & 10.22\,GiB & 22.18 & 0.867 & 0.145 & 85.04 & \qthird{2.46M} & \qsecond{4.57\,GiB} \\
Ortho-3DGS & \qsecond{27.10} & \qsecond{0.872} & \qsecond{0.160} & 54.43 & 5.92M & 9.13\,GiB & \qthird{22.57} & \qsecond{0.853} & \qsecond{0.151} & 74.22 & 5.75M & 9.11\,GiB & \qbest{22.39} & \qsecond{0.869} & \qsecond{0.143} & 63.61 & 3.13M & 5.34\,GiB \\
\cmidrule(lr){1-19}
\textbf{Tortho-Gaussian++} & \qthird{27.05} & \qthird{0.870} & \qthird{0.161} & \qthird{29.70} & \qsecond{2.97M} & \qsecond{6.54\,GiB} & \qsecond{22.59} & 0.844 & 0.167 & \qbest{20.88} & \qsecond{3.06M} & \qsecond{6.70\,GiB} & \qthird{22.21} & 0.859 & 0.158 & \qbest{17.97} & \qsecond{2.28M} & 5.78\,GiB \\
\bottomrule
\end{tabular}%
}
\par\vspace{3pt}
\begin{minipage}{\textwidth}
\scriptsize
$N_{\mathrm{GS}}$ denotes Gaussian number; M denotes $10^6$ Gaussians.
$\uparrow$ and $\downarrow$ indicate that higher and lower values are better, respectively.
\textit{OOM} denotes unsuccessful runs due to out-of-memory errors. \textit{Failed} denotes partitioning failures.
VRAM denotes the reported peak allocated GPU memory; for multi-process methods, it is the maximum per-process value rather than the concurrent total.
\end{minipage}
\endgroup
\end{table*}

\subsection{Qualitative Evaluation}
\label{sec:qualitative}
\begin{figure*}[!t]
    \centering
    \includegraphics[width=0.95\textwidth]{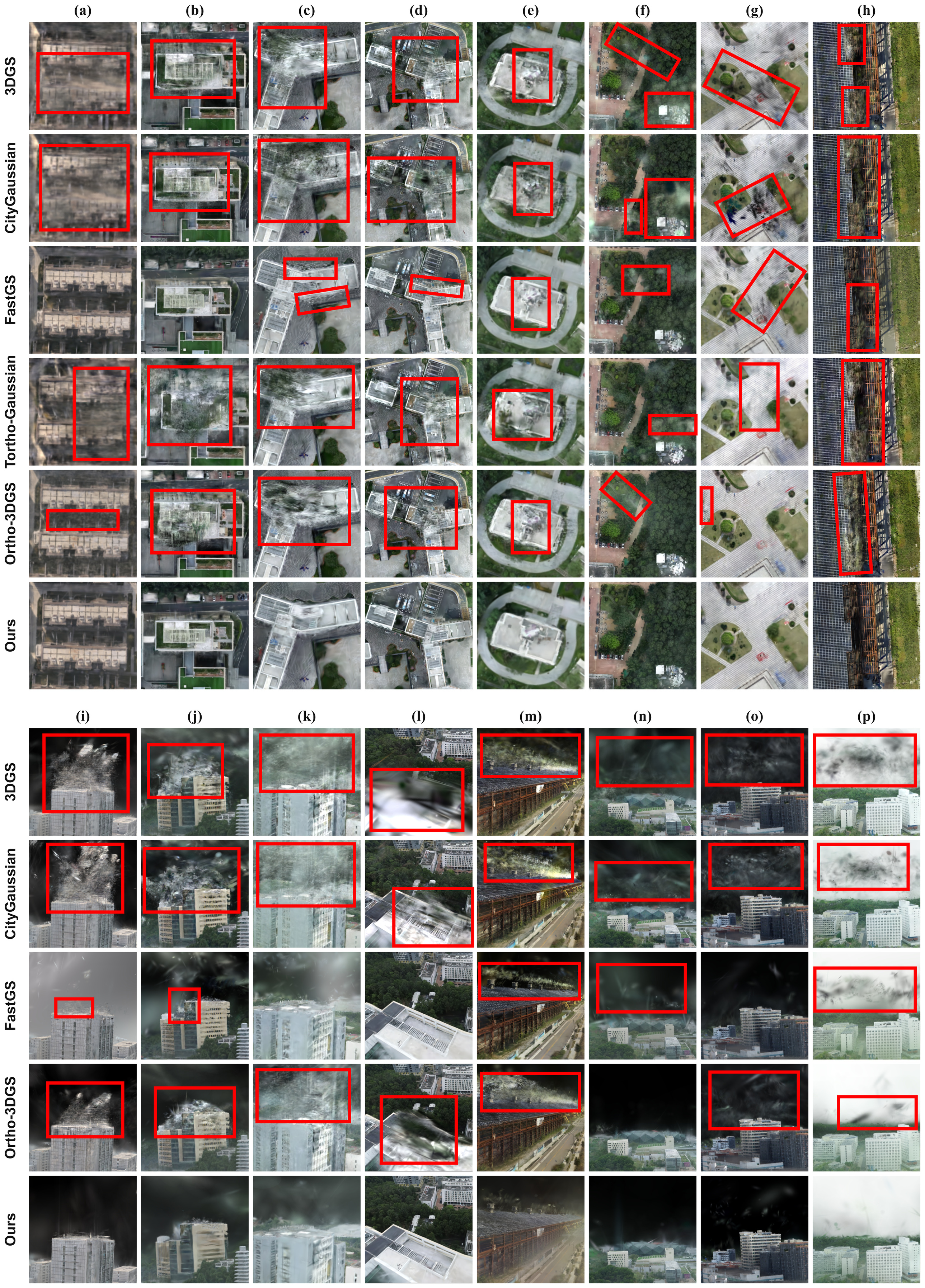}
    \caption{Qualitative comparison of geometric drift and floaters under orthogonal projection (a)--(h) and perspective projection (i)--(p). Red boxes highlight representative structural distortions and spurious patches.}
    \label{fig:drift_floaters}
\end{figure*}

Our qualitative evaluation focuses on geometric drift, floaters, and tilt-induced facade artifacts. For the first two aspects, we compare both general-purpose Gaussian Splatting methods and TDOM-oriented methods across all datasets to examine their impact on TDOM quality. Representative examples are presented under both orthogonal and perspective projection, revealing the appearance of these artifacts in TDOMs and their spatial characteristics. Particularly, since the perspective views are obtained using the default SIBR viewer of vanilla 3DGS, Tortho-Gaussian is omitted due to its modified Gaussian parameterization being incompatible with this viewer (see Fig.~\ref{fig:drift_floaters}). We further compare tilt-induced facade artifacts among the TDOM-oriented methods on the \textit{Tianjin\_Xiqing} and \textit{Hanyang} scenes. Although all compared methods can establish a reasonable projection reference for these scenes, the high-rise buildings impose stringent requirements on the accuracy of the projection direction, allowing us to assess whether Tortho-Gaussian++ maintains competitive TDOM quality under these conditions.

\subsubsection{Geometric Drift and Floaters}
\label{sec:drift}
Fig. \ref{fig:drift_floaters} presents qualitative comparisons of geometric drift and floaters under orthogonal projection (a)–(h) and perspective projection (i)–(p). To ensure a fair comparison, all orthogonal renderings are generated using the same projection reference. The red boxes highlight representative regions where geometric drift and floaters compromise structural clarity and obscure scene content.

In the case of orthogonal projection (i.e., TDOM), geometric drift and floaters manifest as structural distortions and spurious patches. In Fig. \ref{fig:drift_floaters} (a)–(e), competing methods exhibit noticeable blurring or distortion around building boundaries within the red boxes. These artifacts also extend across roof surfaces and adjacent areas, obscuring rooftop details and weakening the separation between buildings and their surroundings. In contrast, Tortho-Gaussian++ substantially suppresses Gaussian drift around buildings, preserving sharp building outlines and well-defined texture details. Furthermore, the red boxes in Fig. \ref{fig:drift_floaters} (f)–(h) highlight spurious patches over vegetation, paved areas, and other structures, which partially obscure the underlying scene content. Tortho-Gaussian++ also removes these visible patches, yielding clearer TDOMs with unobstructed scene details.

The perspective comparisons again reveal the spatial configuration of geometric drift and floaters. In Fig. \ref{fig:drift_floaters} (i)–(m), numerous spurious Gaussians drift upward from building rooftops, approximately along the lines of viewing rays toward the overhead training cameras. This pattern is highly consistent with the possible drift and floater directions predicted by our geometric observability analysis in Fig. \ref{fig:framework} (A). In Fig. \ref{fig:drift_floaters} (n)–(p), all the compared methods exhibit substantial numbers of floating Gaussians above the scene, revealing the three-dimensional form of the spurious patches observed in the orthogonal views. While FastGS and Ortho-3DGS alleviate some of these artifacts, visible residuals remain in several regions. Again, Tortho-Gaussian++ produces significantly fewer and less extensive artifacts, preserving clearer building structures and cleaner scene reconstructions.

Together, the orthogonal and perspective comparisons demonstrate that our Tortho-Gaussian++ can effectively suppress geometric drift and floaters within a prior-free training framework.

\subsubsection{Facade Artifacts} 
\label{sec:facade}
\begin{figure}[pos=!htbp]
    \centering
    \includegraphics[width=\columnwidth]{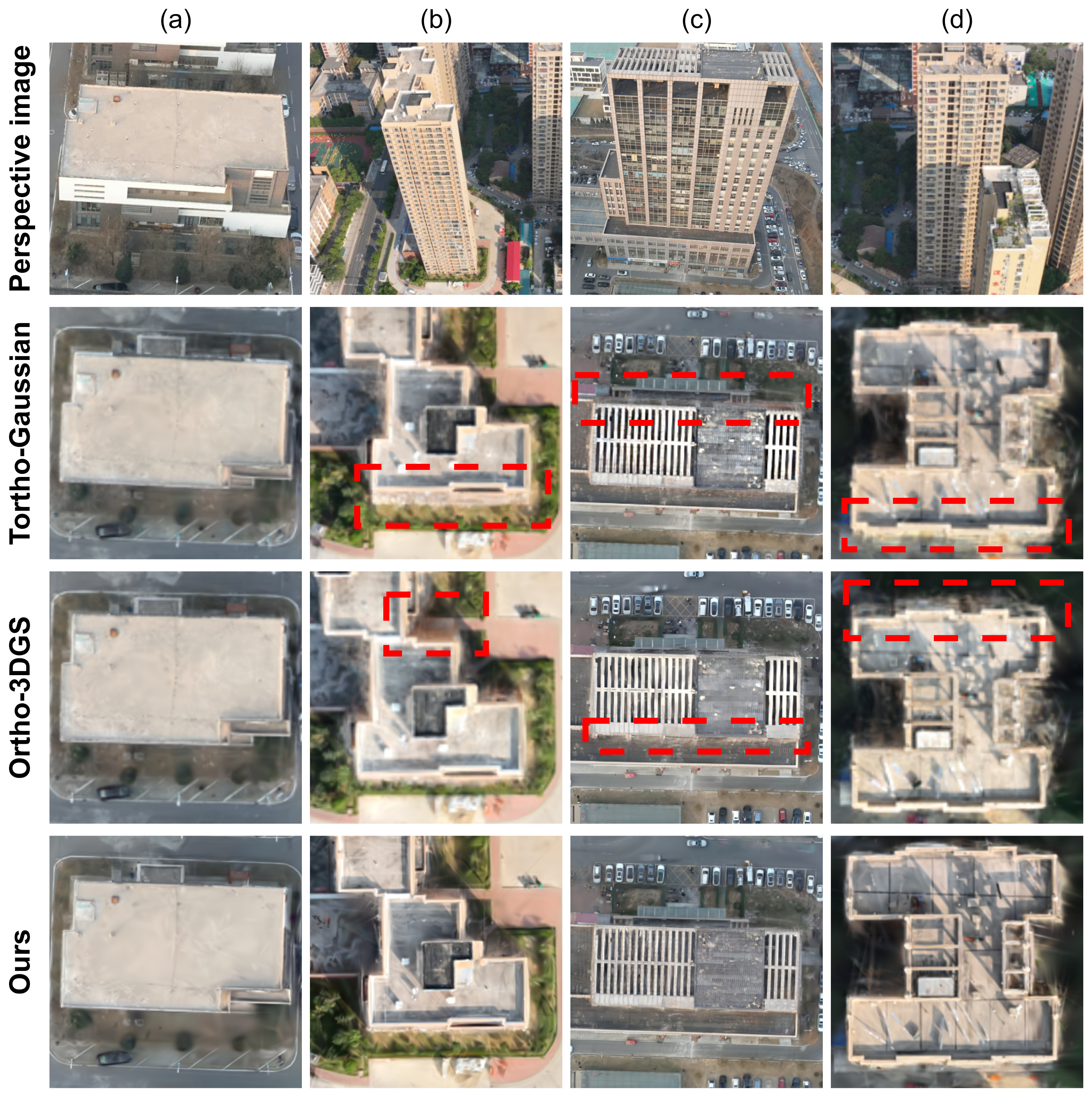}
    \caption{Qualitative comparison of tilt-induced facade artifacts on low-rise (a) and high-rise (b)-(d) buildings. Perspective images illustrate the building heights.}
    \label{fig:facade}
\end{figure}

In an ideal TDOM, building facades should remain invisible. Thus, facade visibility serves as a critical qualitative criterion for assessing TDOM quality. Fig. \ref{fig:facade} presents comparisons of Tortho-Gaussian, Ortho-3DGS, and Tortho-Gaussian++, together with corresponding perspective images that illustrate the building heights. A low-rise building is included as a baseline reference, while the remaining examples focus on high-rise buildings, where even minor deviations in the projection direction can produce conspicuous tilt-induced facade artifacts.

In Fig. \ref{fig:facade} (a), all three methods effectively suppress visible building facades and produce satisfactory orthogonal renderings for the low-rise buildings. However, the limited height of this structure makes facade visibility less sensitive to small errors in the projection direction, providing limited discrimination among the methods.

In contrast, the substantially taller high-rise examples in Fig. \ref{fig:facade} (b)-(d), which remain underexplored in previous studies, reveal pronounced differences. Tortho-Gaussian and Ortho-3DGS exhibit evident tilt-induced facade artifacts along building boundaries, as highlighted by the red boxes. Tortho-Gaussian++ consistently occludes these artifacts, demonstrating substantially less facade visibility across all three examples. Among the compared methods, this more consistent occlusion provides strong qualitative evidence of a more reliable gravity-aligned orthogonal projection reference, even in the scene where all methods can produce generally reasonable orthogonal renderings. This assessment directly complements the robustness analysis under challenging acquisition conditions in Sec. \ref{sec:robustness}.

\subsection{Quantitative Evaluation}
\label{sec:quantitative}
\subsubsection{Rendering Quality}
\label{sec:psnr}
Table \ref{tab:quantitative} compares novel view synthesis quality, time efficiency and resource consumption across twelve UAV scenes. Operating within a prior-free training framework, Tortho-Gaussian++ achieves highly competitive rendering quality. It obtains the highest PSNR on seven scenes and ranks among
the top three on all twelve scenes. Compared with vanilla 3DGS, its mean PSNR increases by 0.95\,dB, with particularly pronounced improvements of 3.97 and 2.94\,dB on \textit{Chengzi} and \textit{DJI}, respectively. It also achieves superior or comparable PSNR to the depth-regulated Ortho-3DGS, outperforming it on seven of the eleven scenes with available results and remaining closely matched on the other four. For example, the gains reach 1.15\,dB on \textit{Chengzi} and 0.55\,dB on \textit{Building}. The modest gains on some scenes may reflect the localized nature of geometric drift and floaters, as improvements from their suppression can be diluted when averaged over many test images. These results should therefore be interpreted alongside the qualitative comparisons in Sec. \ref{sec:qualitative}.

\subsubsection{Time Efficiency and Resource Consumption}
\label{sec:time}
Compared with vanilla 3DGS, Tortho-Gaussian++ substantially reduces mean training time, VRAM usage, and Gaussian number across the twelve scenes by 45.5\%, 47.2\%, and 58.6\%, respectively, as shown in Table \ref{tab:quantitative}. All three resource metrics are reduced on nearly all scenes. The largest reductions reach 69.2\% in training time on CUHK\_UPPER, 61.8\% in VRAM usage on \textit{DJI}, and 74.7\% in Gaussian number on \textit{Tianjin\_Xiqing}. Compared with CityGaussian, Tortho-Gaussian and Ortho-3DGS, which incur additional computational overhead, Tortho-Gaussian++ achieves several-fold training speedups on average while substantially reducing VRAM usage and model size. 

Notably, FastGS achieves shorter training time, with lower VRAM usage and Gaussian number on every scene. However, this aggressive reduction in representation capacity compromises the reconstruction of fine details, as reflected in its lower PSNR and higher LPIPS. Tortho-Gaussian++ achieves higher PSNR and SSIM and lower LPIPS than FastGS on all twelve scenes, with mean improvements of 0.79\,dB, 0.023, and 0.037, respectively. Overall, Tortho-Gaussian++ strikes an optimal balance, maintaining competitive rendering quality while substantially improving time efficiency and reducing resource consumption.

\subsubsection{Planimetric Accuracy}
\label{sec:GCP}
To ensure a rigorous quantitative evaluation, we utilize six ground control points (GCPs) from the \textit{npu} scene to assess the planimetric accuracy of the generated TDOMs. We compare our method against traditional photogrammetric methods (Map2DFusion, ContextCapture, Metashape and Pix4DMapper), general-purpose 3DGS methods (3DGS, FastGS) and TDOM-ASLoriented methods (Tortho-Gaussian, Ortho-3DGS and ours).

GCPs are initially provided as latitude, longitude, and above mean sea level (AMSL), representing orthometric heights. These heights are first converted to ellipsoidal heights using height anomalies from the EGM2008 model. The geographic coordinates are then transformed into a common local east–north–up (ENU) coordinate system through the WGS84 Earth-centered, Earth-fixed (ECEF) coordinate system. To establish a gravity-aligned reference, the ENU coordinate system is rotated using EGM2008-derived deflections of the vertical. Finally, the six GCPs are then orthogonally projected onto the resulting horizontal plane. 

\begin{figure*}[!t]
    \centering
    \includegraphics[width=\textwidth]{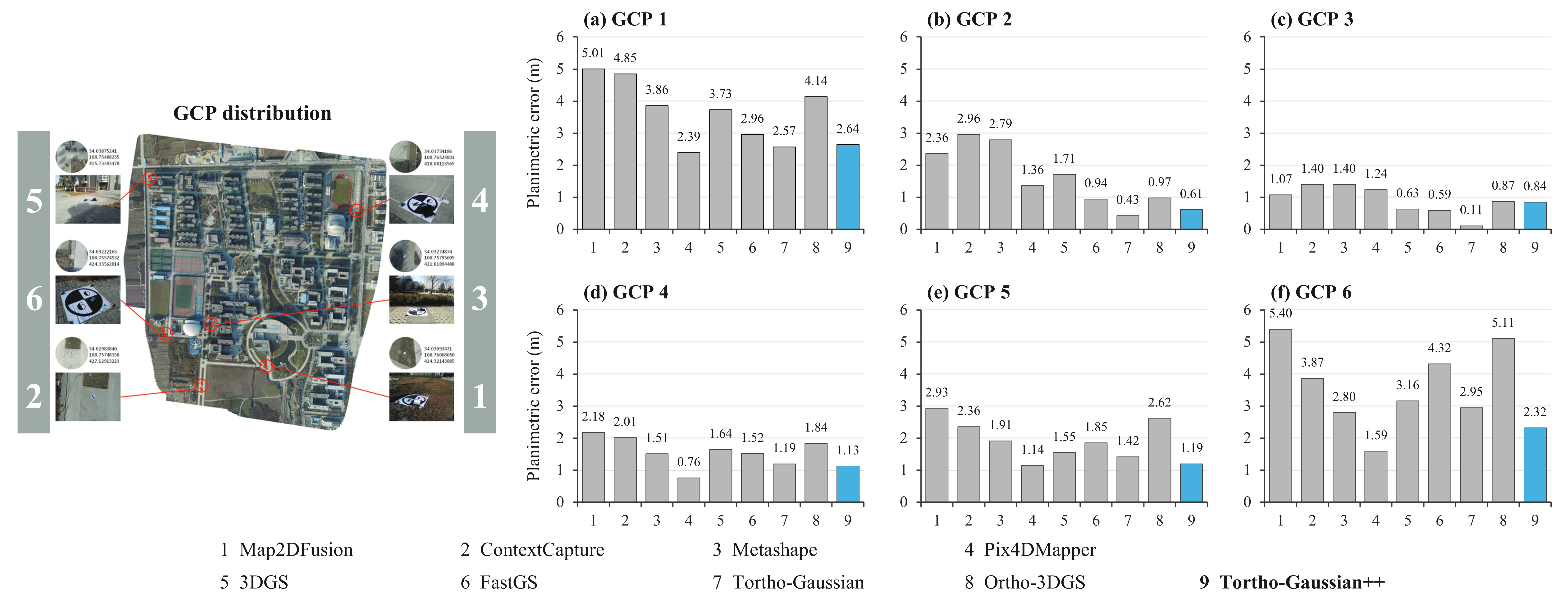}
    \caption{Point-wise planimetric errors at the six GCPs in the \textit{NPU} scene. The top-left image illustrates the distribution of GCPs \citep{bu2016map2dfusion}. Panels (a)--(f) correspond to GCPs~1--6. Blue bars and bold value labels identify Tortho-Gaussian++, while gray bars represent the comparison methods. Lower values indicate smaller post-registration residuals.}
    \label{fig:gcp}
\end{figure*}

For each generated TDOM, the 2D pixel coordinates of the corresponding GCPs are extracted through interactive annotation. The two sets of planar coordinates are registered using a 2D similarity transformation estimated by least-squares adjustment. The planimetric RMSE is calculated over the six points, as summarized in Table \ref{tab:GCP}. Tortho-Gaussian++ ranks second-best among all evaluated methods, achieving planimetric RMSE of 1.645 meters, which is merely 0.143 meters higher than the top-performing commercial software, Pix4DMapper. The point-wise comparison in Fig. \ref{fig:gcp} further reveals the distribution of the remaining planimetric errors. Tortho-Gaussian++ achieves the second-lowest errors at GCP~2, 4, 5, and~6, with values of 0.61, 1.13, 1.19, and 2.32\,m, respectively. It also yields lower errors than Ortho-3DGS at all six GCPs. Although the relative ranking varies across points, these results complement the aggregate RMSE and demonstrate competitive post-registration planimetric accuracy.

\begin{table*}[!t]
\centering
\captionsetup{format=plain,justification=raggedright,
  singlelinecheck=false,labelsep=space,margin=0pt}
\caption{Result of planimetric RMSE with six GCPs in the \textit{npu} scene.}
\label{tab:GCP}
\begingroup
\small
\setlength{\tabcolsep}{4pt}
\renewcommand{\arraystretch}{1.15}

\resizebox{\textwidth}{!}{%
\begin{tabular}{@{}l*{9}{c}@{}}
\toprule
Method
& Map2DFusion & ContextCapture & Metashape & Pix4DMapper
& 3DGS & FastGS & Tortho-Gaussian & Ortho-3DGS & Tortho-Gaussian++ \\
\midrule
\textbf{RMSE (m)}$\downarrow$
& 3.518 & 3.130 & 2.530 & \textbf{1.502}
& 2.321 & 2.393 & 1.776 & 3.032 & \underline{1.638} \\
\bottomrule
\end{tabular}%
}

\par\vspace{3pt}
\begin{minipage}{\textwidth}
\footnotesize
$\downarrow$ indicates that lower values are better.
The best and second-best results are highlighted in bold and underlined, respectively.
\end{minipage}
\endgroup
\end{table*}

\subsection{Robustness of Projection Reference}
\label{sec:robustness}
While the facade comparisons in Sec. \ref{sec:qualitative} examine the reliability of the projection reference where competing methods yield reasonable renderings, this section further investigates its robustness under challenging image acquisition conditions using the \textit{SMBU}, \textit{UPPER}, and \textit{Residence} datasets.

The \textit{SMBU} and \textit{UPPER} scenes are composed of approximately 80\% oblique imagery, which severely limits the availability for methods that manually rotate nadir cameras to define the projection reference. Furthermore, \textit{Residence} consists entirely of oblique imagery and exhibits substantial variations in flight altitude. Such altitude fluctuations cause methods \citep{yang2025ortho} that use a plane fitted to the camera centers as a horizontal reference to fail, as the normal of the fitted plane deviates considerably from the true gravity direction, thereby violating the constant-altitude prior.

\begin{figure}[pos=!htbp]
    \centering
    \includegraphics[width=\columnwidth]{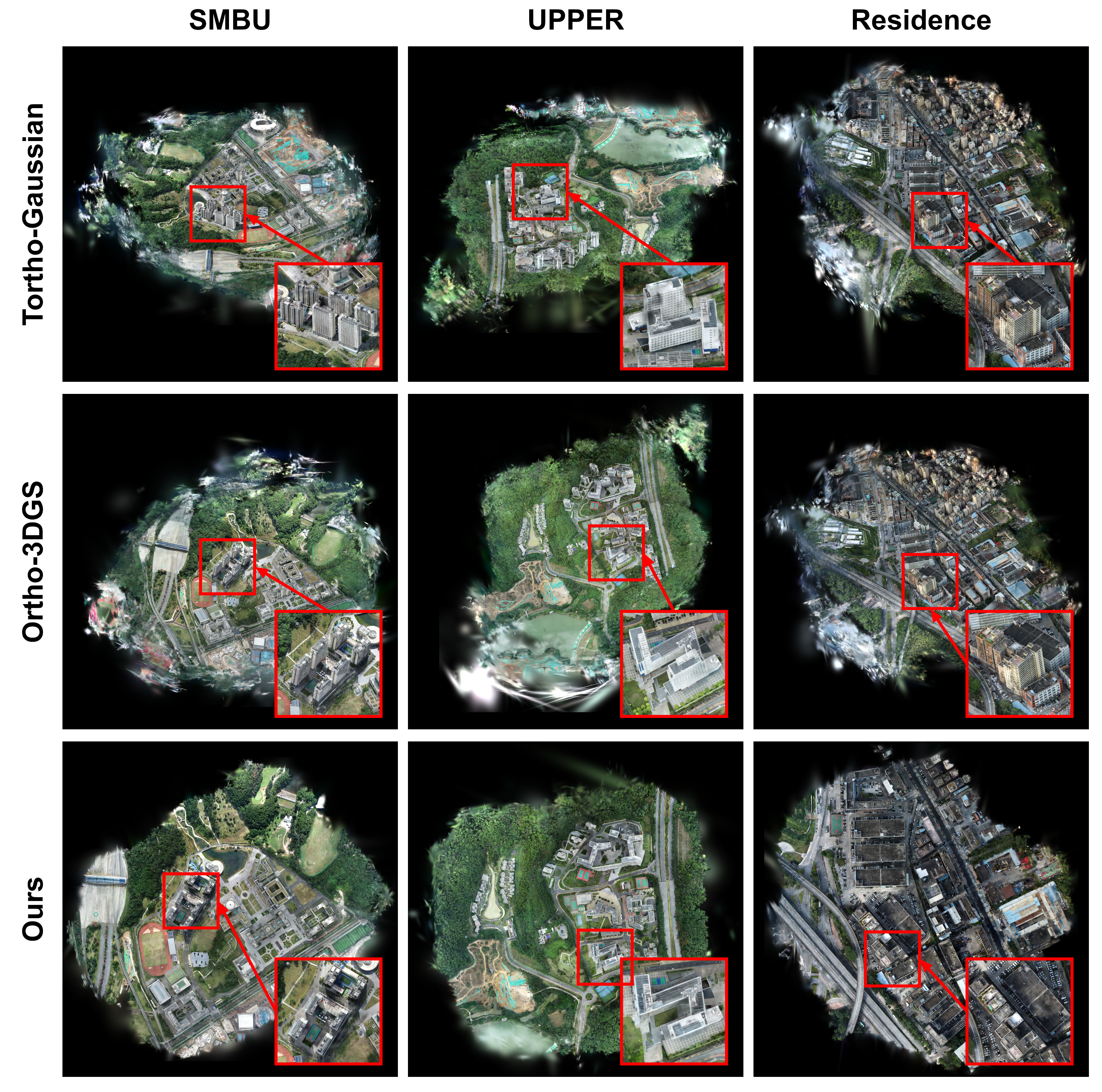}
    \caption{Qualitative comparison of TDOMs under challenging acquisition conditions. \textit{SMBU} and \textit{UPPER} are composed of 80\% oblique imagery, while \textit{Residence} consists entirely of oblique views with substantial flight-altitude variations. The proposed method maintains a gravity-aligned projection, effectively suppressing tilt-induced facade artifacts where competing methods fail.}
    \label{fig:robustness}
\end{figure}

As shown in Fig. \ref{fig:robustness}, Tortho-Gaussian and Ortho-3DGS yield projection references that are severely skewed across all three challenging scenes. Consequently, the resulting renderings expose extensive building facades, fundamentally compromising the strict orthogonal nature required for TDOMs. In stark contrast, Tortho-Gaussian++ maintains a highly consistent, gravity-aligned projection reference with substantially reduced facade visibility. These results demonstrate the robustness of the proposed multi-source gravity direction estimation framework under challenging and complex UAV image acquisition conditions.

\begin{figure*}[!t]
    \centering
    \includegraphics[width=\textwidth]{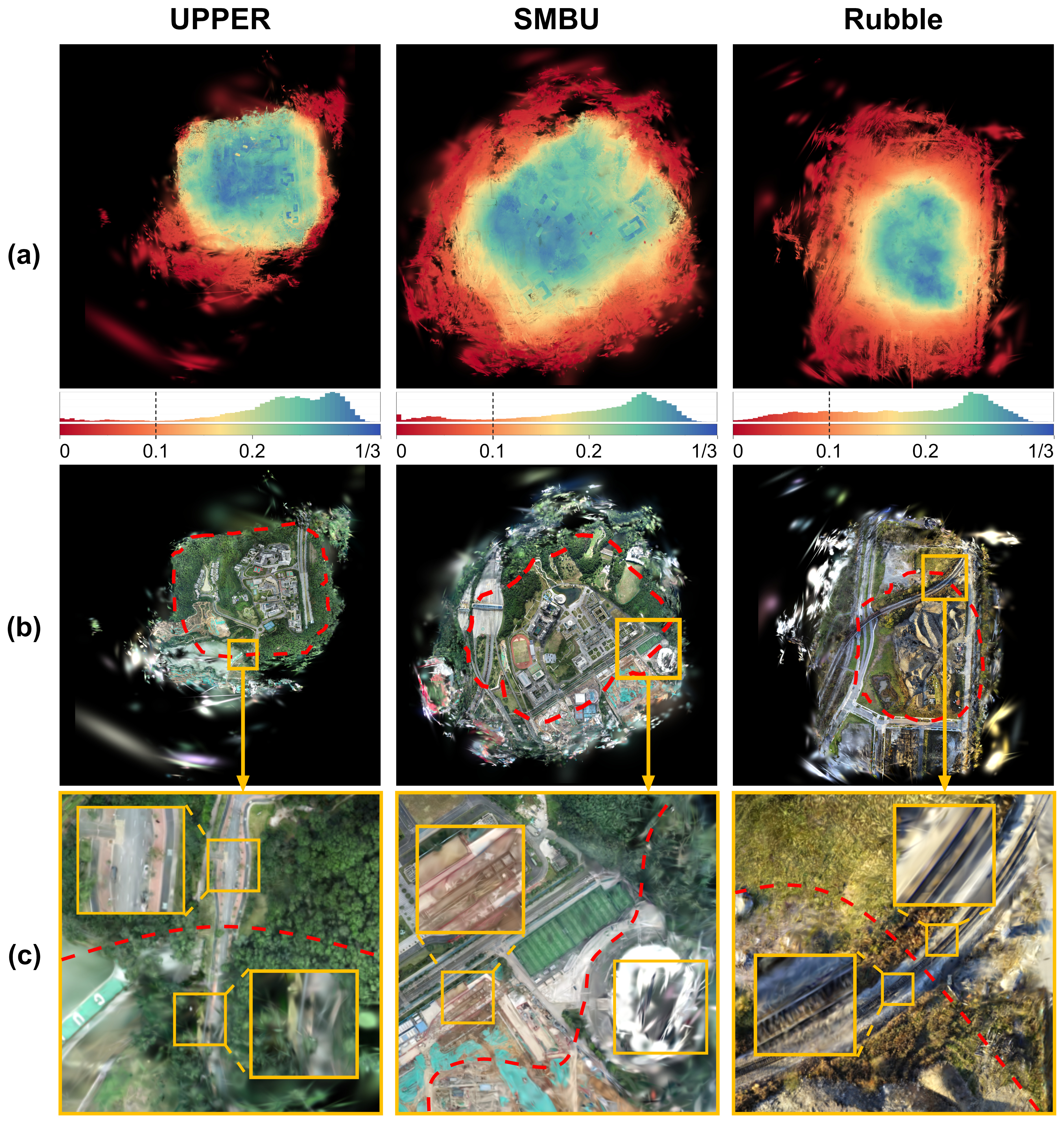}
    \caption{Geometric support-based pruning on \textit{UPPER}, \textit{SMBU}, and \textit{Rubble}. (a) Spatial distributions and histograms of geometric support scores, with red and blue indicating low and high support, respectively. Black dashed lines mark the pruning threshold $\tau_s=0.1$. (b) TDOMs with red dashed contours outlining the retained regions. (c) Enlarged views of the yellow-boxed regions, comparing reconstruction quality immediately on either side of the pruning boundaries.}
    \label{fig:prune}
\end{figure*}

\subsection{Ablation Studies}
\label{sec:ablation}
In this section, extensive ablation studies are conducted to evaluate the performance of the proposed components: geometric support-based pruning, optimizer-state inheritance, and robust gravity direction estimation. These experiments systematically examine the identification and removal of unreliable peripheral regions, the effects of optimizer-state inheritance on rendering quality, time efficiency and resource consumption, and the complementary roles of image-derived and camera-distribution-based gravity estimates.

\subsubsection{Effect of Geometric Support-Based Pruning}
\label{sec:ablation1}
\textbf{Pruning effectiveness.} We first evaluate the efficacy of the geometric support score as a criterion for identifying unreliable peripheral reconstructions. Fig. \ref{fig:prune} presents the spatial distribution of geometric support scores, the pruning boundaries, and local details for the \textit{UPPER}, \textit{SMBU}, and \textit{Rubble} scenes. Low scores predominantly occur around scene boundaries, where the reconstructed regions exhibit severe stretching, blurring, and isolated floaters. As shown in Fig. \ref{fig:prune} (a), interior regions consistently yield higher scores and more coherent structures. This spatial correspondence indicates that the proposed viewing-ray-based geometric observability analysis can faithfully capture variations in reconstruction reliability.

The red dashed contours in Fig. \ref{fig:prune} (b) delineate the boundaries of the retained regions. The enlarged views in Fig. \ref{fig:prune} (c) further compare reconstruction quality immediately on either side of these boundaries. Roads, buildings, and railways remain structurally intact with clear textures on the retained side, while the excluded regions exhibit substantial distortion or loss of texture detail. These observations demonstrate that our automatic pruning mechanism accurately delineates boundaries, effectively removing poorly reconstructed peripheral regions while preserving high-fidelity interior structures.

\begin{figure*}[!t]
    \centering
    \includegraphics[width=\textwidth]{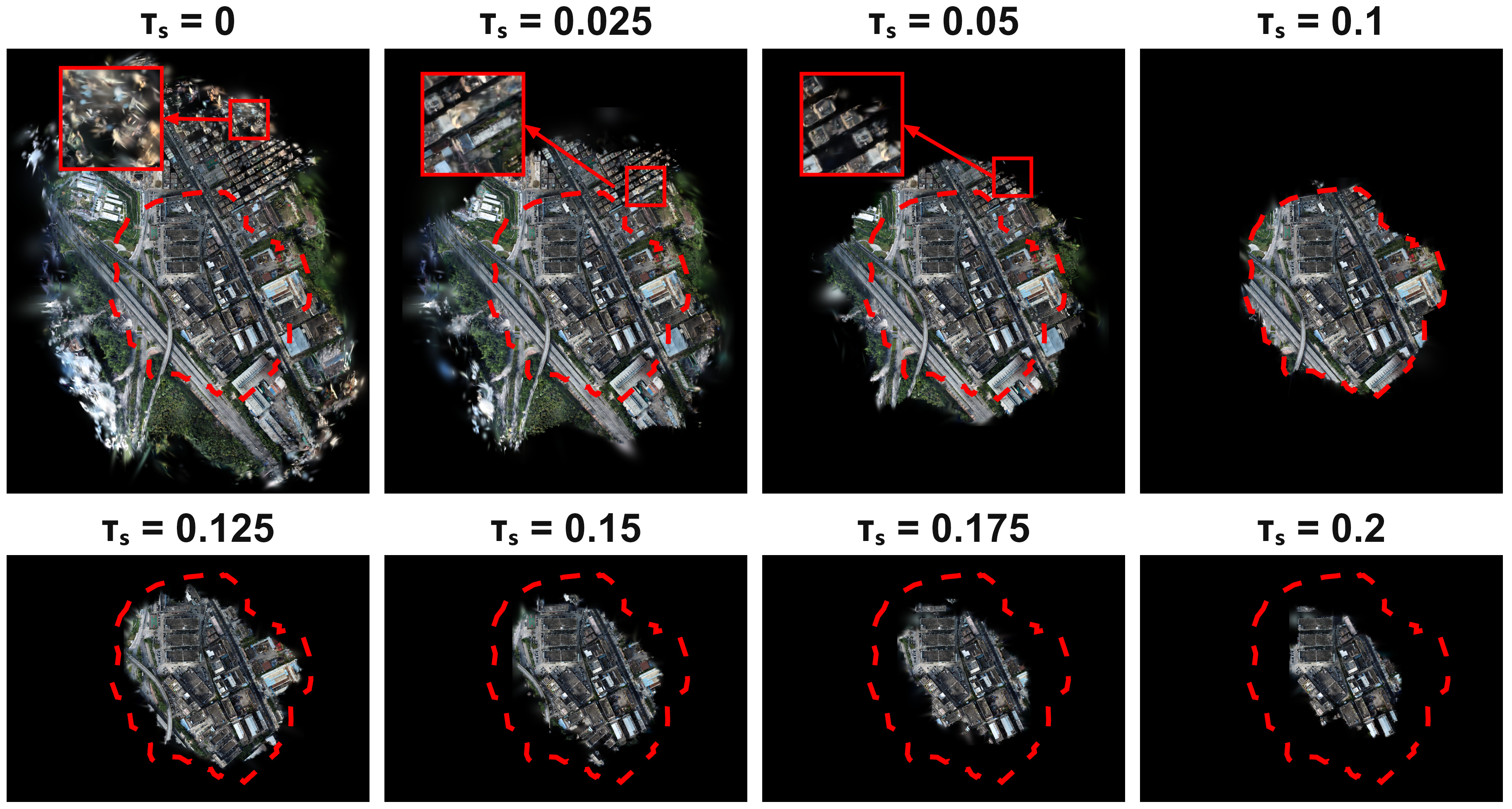}
    \caption{Sensitivity analysis of the pruning threshold $\tau_s$ on the \textit{Residence} scene. All main views are displayed at the same spatial scale. Red dashed contours mark the boundary retained at $\tau_s=0.1$.}
    \label{fig:tau}
\end{figure*}

\textbf{Threshold sensitivity.} We further investigate the sensitivity to the pruning threshold $\tau_s$ using the \textit{Residence} scene. As shown in Fig. \ref{fig:tau}, we evaluate eight values of the threshold $\tau_s$: 0, 0.025, 0.05, 0.1, 0.125, 0.15, 0.175, and 0.2,
where $\tau_s=0$ serves as the reference without support-based pruning. Small thresholds remove minor peripheral artifacts but fail to eliminate visibly distorted reconstructions. At $\tau_s=0.1$, the majority of peripheral clutter is removed while the major interior structures remain intact. Further increases, particularly towards $0.175$ and $0.2$, aggressively exclude additional peripheral structures and significantly reduce scene coverage. The comparison illustrates the inherent trade-off between removing unreliable reconstructions and retaining spatial coverage, justifying the empirical choice of $\tau_s=0.1$ used in our experiments.

\subsubsection{Effect of Optimizer-State Inheritance}
\label{sec:ablation2}
We evaluate optimizer-state inheritance (OSI) on \textit{Chengzi}, \textit{DJI}, and \textit{CUHK\_LOWER} using four configurations: vanilla 3DGS, 3DGS with geometric observability analysis (w/ GOA), 3DGS with optimizer-state inheritance (w/ OSI), and Full, which combines GOA and OSI. To evaluate the optimization components independently, the proposed automatic pruning is disabled for all configurations, including Full. Table \ref{tab:Inheritance} reports rendering quality, training time, Gaussian number, and VRAM usage.

Integrating OSI into vanilla 3DGS increases PSNR by 3.14, 2.93, and 0.67\,dB on \textit{Chengzi}, \textit{DJI}, and \textit{CUHK\_LOWER}, respectively, with corresponding gains in SSIM, although LPIPS rises slightly on some scenes. In addition, w/ OSI reduces training time by approximately 46.9--57.1\%, Gaussian number by 51.3--62.2\%, and VRAM usage by 54.1--64.8\%. Since automatic pruning is disabled here, these results indicate that OSI alone yields more compact Gaussian representations during training, lowering resource consumption while improving PSNR. By contrast, w/ GOA alone provides negligible improvements in rendering quality, as its primary function is to quantify geometric support and establish the criterion for subsequent pruning. The Full configuration attains a slightly higher PSNR than w/ OSI, albeit with a slight increase in training time and VRAM. This marginal overhead is consistent with the computational cost of maintaining geometric support tensors during optimization. Collectively, these results clearly delineate the distinct contributions of the two components: OSI is primarily driver for enhancing rendering quality, training efficiency and resource consumption, whereas GOA is essential for identifying and pruning unreliable peripheral regions.

\begin{table}[htbp]
\centering
\captionsetup{format=plain,justification=raggedright,
  singlelinecheck=false,labelsep=newline,margin=0pt}
\caption{Ablation study of geometric observability analysis (GOA) and optimizer-state inheritance (OSI) on three scenes, with automatic pruning disabled for all methods. Full combines both components. The best results within each scene are highlighted in bold.}
\label{tab:Inheritance}
\begingroup
\footnotesize
\setlength{\tabcolsep}{2pt}
\renewcommand{\arraystretch}{1.18}

\resizebox{\columnwidth}{!}{%
\begin{tabular}{@{}clcccccc@{}}
\toprule
Scene & \multicolumn{1}{c}{Method}
& PSNR$\uparrow$
& SSIM$\uparrow$
& LPIPS$\downarrow$
& Time$\downarrow$
& $N_{\mathrm{GS}}\downarrow$
& VRAM$\downarrow$ \\
\midrule

\multirow{4}{*}{Chengzi}
 & 3DGS
 & 25.04 & 0.874 & 0.157
 & 53.00 & 14.63M & 23.03\,GiB \\
 & w/ GOA
 & 25.29 & 0.874 & \textbf{0.157}
 & 51.60 & 14.47M & 22.78\,GiB \\
 & w/ OSI
 & 28.18 & 0.900 & 0.159
 & \textbf{28.12} & \textbf{7.13M} & \textbf{10.58\,GiB} \\
 & \textbf{Full}
 & \textbf{28.35} & \textbf{0.901} & 0.158
 & 33.80 & 7.15M & 11.40\,GiB \\
\midrule

\multirow{4}{*}{DJI}
 & 3DGS
 & 24.54 & 0.863 & 0.104
 & 62.38 & 17.89M & 29.07\,GiB \\
 & w/ GOA
 & 24.38 & 0.860 & 0.107
 & 63.52 & 17.80M & 31.24\,GiB \\
 & w/ OSI
 & 27.47 & 0.914 & \textbf{0.083}
 & \textbf{26.78} & 6.91M & \textbf{10.23\,GiB} \\
 & \textbf{Full}
 & \textbf{27.49} & \textbf{0.914} & \textbf{0.083}
 & 27.67 & \textbf{6.87M} & 11.12\,GiB \\
\midrule

\multirow{4}{*}{CUHK\_LOWER}
 & 3DGS
 & 27.56 & 0.892 & \textbf{0.128}
 & 32.96 & 6.96M & 10.64\,GiB \\
 & w/ GOA
 & 27.46 & 0.889 & 0.130
 & 35.23 & 6.98M & 12.86\,GiB \\
 & w/ OSI
 & 28.23 & 0.899 & 0.140
 & \textbf{15.23} & \textbf{2.63M} & \textbf{4.03\,GiB} \\
 & \textbf{Full}
 & \textbf{28.40} & \textbf{0.899} & 0.140
 & 16.25 & \textbf{2.63M} & 6.07\,GiB \\
\bottomrule
\end{tabular}%
}

\par\vspace{3pt}
\begin{minipage}{\columnwidth}
\footnotesize
The w/ GOA and w/ OSI methods add the corresponding component to vanilla 3DGS.
$N_{\mathrm{GS}}$ denotes Gaussian number; M denotes $10^6$ Gaussians.
$\uparrow$ and $\downarrow$ indicate that higher and lower values are better, respectively.
\end{minipage}
\endgroup
\end{table}

\subsubsection{Effect of Gravity Direction Estimation}
\label{sec:ablation3}
\begin{figure}[pos=!htbp]
    \centering
    \includegraphics[width=\columnwidth]{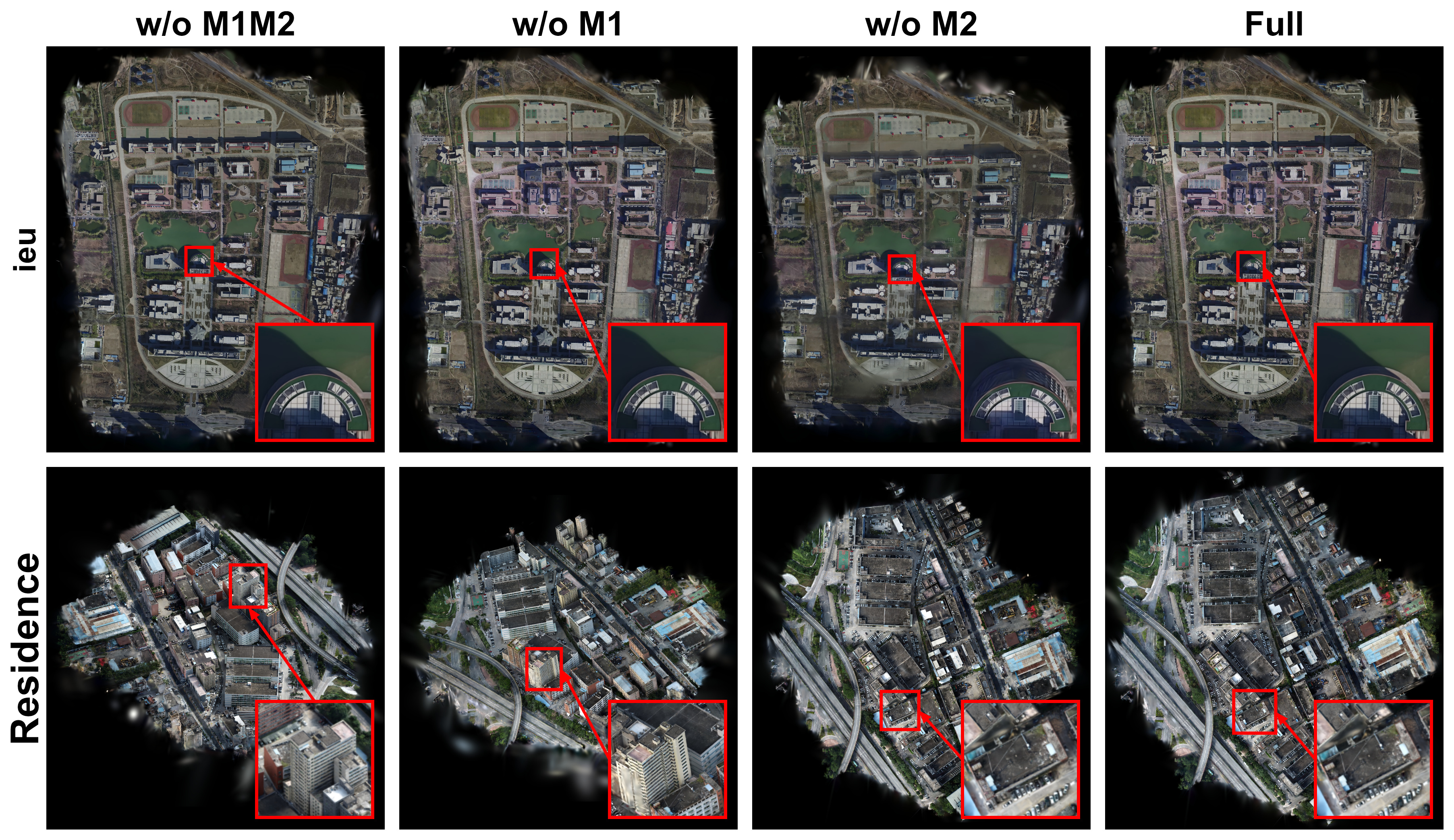}
    \caption{Ablation study of gravity direction estimation on \textit{ieu} and \textit{Residence}. M1 and M2 denote image-derived and camera-distribution-based gravity estimation, respectively. Full uses the proposed confidence-based selection and fusion strategy, while w/o M1M2 adopts an arbitrary camera orientation as the projection reference. Enlarged regions highlight differences in tilt-induced facade artifacts.}
    \label{fig:ablationMM}
\end{figure}

We investigate the individual and combined contributions of image-derived gravity direction estimation (M1) and camera-distribution-based gravity direction estimation (M2) using the \textit{ieu} and \textit{Residence} scenes. Fig. \ref{fig:ablationMM} compares four configurations: w/o M1M2, w/o M1, w/o M2, and Full. When both M1 and M2 are disabled (w/o M1M2), the orthogonal projection reference is determined by the orientation of an arbitrarily selected camera. 

The \textit{ieu} scene is a nadir-only dataset dominated by low-rise buildings. The limited visibility of vertical building edges provides insufficient cues for the single-image model used in M1, resulting in an unreliable estimation. Conversely, the stable flight altitude allows M2 to reliably estimate the normal to the plane fitted to the camera centers. Consequently, the w/o M2 configuration introduces noticeable tilt-induced facade artifacts, as illustrated by the enlarged region in Fig. \ref{fig:ablationMM}. Retaining M2 produces a top-down appearance close to that of Full, indicating that the camera-distribution-based estimate provides a highly effective projection reference for scenes with stable flight altitudes. 

\textit{Residence}, however, exhibits substantial flight-altitude variations that invalidate the constant-altitude prior underlying M2. Moreover, its abundant oblique views provide rich informative perspective cues, allowing the model in M1 to estimate the gravity direction effectively. Accordingly, the w/o M1M2 and w/o M1 configurations produce markedly oblique renderings with extensive facade visibility, whereas retaining M1 in w/o M2 and Full substantially reduces tilt-induced facade artifacts. These results demonstrate the critical importance of image-derived gravity cues under complex acquisition conditions.

The two scenes exhibit distinct dependencies on the gravity information sources: M2 provides an effective constraint under stable altitude conditions, whereas M1 is particularly important on oblique imagery. The Full configuration consistently minimizes tilt-induced facade artifacts across both scenes. These qualitative results validate the complementary roles of the two gravity estimation modules, demonstrating that fusing multiple information sources is essential for establishing a reliable orthogonal projection reference across diverse and challenging acquisition conditions.

\subsection{Integration into 3DGS-based Pipelines}
\label{sec:integration}

The proposed optimizer-state inheritance strategy functions as a modular, plug-and-play component that can be seamlessly integrated into existing Gaussian Splatting-based pipelines to improve rendering quality, training efficiency and resource consumption. To demonstrate its generalizability beyond vanilla 3DGS, we incorporate our OSI strategy into two representative variants, 2DGS \citep{huang20242d} and PGSR \citep{chen2024pgsr}, and evaluate each baseline against its enhanced variant (+ Ours) on the \textit{UPPER} dataset. All variants are trained for 30,000 iterations to ensure a fair comparison.

As summarized in Table \ref{tab:integration}, integrating OSI yields consistent performance gains across both architectures. Specifically, it improves PSNR by 0.44 and 2.93\,dB for 2DGS and PGSR, respectively, accompanied by corresponding increases in SSIM. Furthermore, the enhanced variants achieve substantial reductions in computational overhead: training time is reduced by 16.6\% and 5.7\%, Gaussian number is decreased by 22.8\% and 15.7\%, and peak VRAM usage drops by 23.1\% and 23.2\% for 2DGS and PGSR, respectively. Regarding perceptual quality, LPIPS shows a notable improvement for PGSR, while a marginal increase is observed for 2DGS, which is a common trade-off when optimizing for structural fidelity (PSNR/SSIM) and model compactness. Overall, these results validate the generalizability of the proposed strategy, confirming that it effectively mitigates the underlying optimization misalignment across various Gaussian Splatting architectures, thereby producing more compact representations with lower training costs and competitive rendering quality.

\begin{table}[!t]
\centering
\captionsetup{
  format=plain,
  justification=raggedright,
  singlelinecheck=false,
  labelsep=newline,
  margin=0pt
}
\caption{Plug-and-play integration of the proposed training strategy into 2DGS and PGSR on \textit{UPPER}. + Ours denotes each method enhanced with our strategy. All variants are trained for 30,000 iterations. The better result within each baseline--variant pair is highlighted in bold.}
\label{tab:integration}

\begingroup
\footnotesize
\setlength{\tabcolsep}{3pt}
\renewcommand{\arraystretch}{1.18}

\resizebox{\columnwidth}{!}{%
\begin{tabular}{@{}lcccccc@{}}
\toprule
Method
& PSNR$\uparrow$
& SSIM$\uparrow$
& LPIPS$\downarrow$
& Time$\downarrow$
& $N_{\mathrm{GS}}\downarrow$
& VRAM$\downarrow$ \\
\midrule

2DGS
& 25.67
& 0.822
& \textbf{0.197}
& 18.05
& 2.71M
& 4.25\,GiB \\

2DGS + Ours
& \textbf{26.11}
& \textbf{0.826}
& 0.205
& \textbf{15.05}
& \textbf{2.09M}
& \textbf{3.26\,GiB} \\

\midrule

PGSR
& 22.75
& 0.694
& 0.265
& 105.60
& 5.97M
& 17.37\,GiB \\

PGSR + Ours
& \textbf{25.67}
& \textbf{0.835}
& \textbf{0.157}
& \textbf{99.56}
& \textbf{5.03M}
& \textbf{13.33\,GiB} \\

\bottomrule
\end{tabular}%
}

\par\vspace{3pt}
\begin{minipage}{\columnwidth}
\footnotesize
$N_{\mathrm{GS}}$ denotes Gaussian number; M denotes $10^6$ Gaussians.
$\uparrow$ and $\downarrow$ indicate that higher and lower values are better, respectively.
\end{minipage}
\endgroup
\end{table}

\section{Discussion}
\label{sec:discussion}
The experimental evaluations not only demonstrate the effectiveness of Tortho-Gaussian++ but also provide critical insights into the fundamental challenges of extending Gaussian splatting from novel view synthesis to reliable 2D photogrammetric mapping.

\textbf{Mechanisms and suppression of geometric drift and floaters.} Our findings provide an optimization-based explanation for geometric drift and floaters in UAV scenes. Specifically, these artifacts manifest as Gaussian displacements along weakly constrained viewing-ray directions during 3DGS training, which can occur even with reasonably accurate SfM initialization. We identify three interacting factors driving this behavior. First, UAV observations (whether oblique or nadir) often exhibit limited viewing-ray diversity, leaving Gaussian positions weakly constrained along specific directions and creating vulnerable conditions for geometric drift and floaters. Second, the \textit{moment-step misalignment} during densification amplifies the parameter updates of newly created Gaussians, providing a driving mechanism for their displacement. Third, in large-scale UAV scenes, an individual Gaussian is typically visible in only a small subset of training images; yet, synchronous optimization allows these amplified updates to continue even when the Gaussian is invisible. Collectively, these factors promote the accumulation of Gaussian displacement. While additional geometric priors can alleviate the resulting artifacts, they do not address the intrinsic causes of geometric drift and floaters. Moreover, \cite{kingma2014adam} show that gradient coupling in Adam can destabilize the effective strength of regularization. Coupled with the residual artifacts observed in depth-guided methods like Ortho-3DGS \citep{yang2025ortho}, this suggests that merely introducing external priors may be insufficient to fully suppress drift and floaters in UAV scenes.

Optimizer-state inheritance directly addresses this mechanism by aligning the moment history with the optimizer step count. This mitigates the amplified updates that drive geometric drift and intrinsically suppresses these artifacts without relying on external geometric priors. Qualitative comparisons show substantially reduced or eliminated drift and floaters, while quantitative results demonstrate additional PSNR gains and reduced resource consumption. Furthermore, integration experiments with 2DGS and PGSR validate that this strategy can seamlessly benefit different Gaussian splatting pipelines through a simple modification to optimizer-state handling during densification.

\textbf{Quantitatively characterize geometric support for individual Gaussians.} Our geometric observability analysis provides an explicit and quantitative characterization of the directional constraints on individual Gaussians. The strong correspondence between low support scores and unreliable peripheral reconstructions in Sec. \ref{sec:ablation1} confirms that the score effectively captures the degree of geometric constraint on individual Gaussians. Beyond automatic pruning, this per-Gaussian reliability characterization holds significant potential for downstream applications, such as uncertainty-aware urban analysis or adaptive level-of-detail modeling, where distinguishing well-constrained regions from geometrically ambiguous ones is crucial.

\textbf{Reliable projection reference for TDOM generation.} The experimental results underscore that reliable scene geometry must be paired with a rigorously defined orthogonal projection direction. Existing methods are often tailored to specific, idealized acquisition conditions (e.g., constant flight altitude or abundant nadir views), which limits their across-dataset applicability. Our robust gravity direction estimation, by contrast, adaptively fuses multiple cues, making it particularly effective under diverse acquisition conditions. This ensures a highly reliable orthogonal projection reference, which is a strict prerequisite for generating artifact-free TDOMs.

\textbf{Limitations and future work.} Despite these improvements, several limitations remain to be addressed.

First, for nadir-only datasets, most Gaussians receive geometric support scores below the empirical pruning threshold, indicating insufficient directional constraints under our current scalar observability metric. This occurs because nearly parallel viewing rays provide weak constraints in the vertical plane. However, for TDOM generation, which primarily relies on horizontal geometric constraints, this vertical weakness has minimal impact on the final orthogonal rendering. Therefore, applying a uniform scalar threshold in such cases leads to over-pruning of valid regions. For this reason, we did not apply geometric support-based pruning to nadir-only datasets. This highlights a fundamental limitation of isotropic scalar evaluation. Future work could exploit the full directional information retained in the geometric support tensor, rather than relying solely on its minimum eigenvalue. As defined in Eq. \ref{dfd}, \(\mathbf d^\top\mathbf F_n\mathbf d\) quantifies the geometric support along direction \(\mathbf d\). Evaluating these directional constraints against the target projection direction would enable a view-dependent assessment, accommodating the distinct geometric requirements of nadir versus oblique rendering.

Second, while optimizer-state inheritance effectively suppresses geometric drift and floaters, inheriting the parent's accumulated moment estimates inherently restricts the exploratory capacity of newly created Gaussians. This trade-off is reflected in the marginally higher LPIPS values observed on certain scenes, as the model may sacrifice the exploration of extremely fine, high-frequency textures in favor of structural stability. Future iterations could explore adaptive moment scaling to balance drift suppression with high-frequency detail recovery.

Third, our optimizer-state inheritance strategy acts as a stabilizer rather than a corrector. It is designed to prevent Gaussians from drifting excessively away from their initial geometric configuration, but it cannot recover geometry from fundamentally flawed initializations. When poor SfM initialization places Gaussians at highly inaccurate positions, common in textureless and highly repetitive or unreliable pheripheral regions, our approach cannot reliably correct these gross topological errors. While our pruning mechanism effectively masks these artifacts, improving initialization quality (e.g., via robust feature matching or prior-guided SfM) remains essential for fully reconstructing such challenging regions.

Finally, the proposed gravity estimation approximates the scene's vertical axis using visual cues and camera-plane priors, rather than deriving the true geodetic gravity vector through rigorous gravimetric calculations (e.g., incorporating local geoid models or precise IMU data). Consequently, while the method provides a highly robust estimate for establishing the orthogonal projection reference in TDOM, it does not guarantee absolute agreement with the true local gravity direction in a strict geodetic sense. Integrating precise inertial or gravimetric data could further elevate the absolute geodetic accuracy of the projection reference in future applications.

\section{Conclusion}
\label{sec:conclusion}
In this paper, we propose Tortho-Gaussian++, a novel framework for high-fidelity true digital orthophoto map (TDOM) generation in large-scale aerial scenes. Concretely, Tortho-Gaussian++ introduces viewing-ray-based geometric observability analysis to quantitatively characterize per-Gaussian support and prune unreliable peripheral regions. We also develop an optimizer-state inheritance strategy to resolve \textit{moment-step misalignment} during densification, intrinsically suppressing geometric drift and floaters without relying on external geometric priors. Furthermore, we design a confidence-based fusion mechanism that integrates image-derived cues, camera-distribution priors, and nadir views to establish a highly reliable, gravity-aligned orthogonal projection reference.

Extensive experiments across twelve UAV scenes demonstrate that Tortho-Gaussian++ achieves competitive rendering quality, improving the average PSNR by 0.95\,dB over vanilla 3DGS while reducing mean training time, VRAM usage, and Gaussian count by 45.5\%, 47.2\%, and 58.6\%, respectively. Qualitative evaluations confirm the substantial suppression of drift and floaters, yielding sharper building outlines and significantly fewer tilt-induced facade artifacts, even under challenging acquisition conditions. Moreover, the seamless integration of our optimizer-state inheritance into 2DGS and PGSR validates its broad applicability across diverse Gaussian splatting pipelines. Ultimately, our Tortho-Gaussian++ provides a highly efficient, prior-free, and practical solution for generating high-quality TDOMs from UAV imagery.



\printcredits

\section*{Acknowledgements}
This work was supported by the National Natural Science Foundation of China under Grant 42671570 awarded to Zongqian Zhan, and by the National Natural Science Foundation of China under Grant 42301507 awarded to Xin Wang.
\bibliographystyle{cas-model2-names}

\bibliography{references}




\end{document}